\documentclass[11pt,letterpaper]{article}

\usepackage[english]{babel}
\usepackage{iftex}
\ifPDFTeX
  \usepackage[T1]{fontenc}
  \usepackage[utf8]{inputenc}
  \usepackage{lmodern}
\else
  \usepackage{fontspec}
\fi
\usepackage{microtype}

\usepackage[margin=1in]{geometry}
\usepackage{setspace}

\usepackage{amsmath, amsthm, amssymb, amsfonts}

\usepackage{algorithm}

\usepackage{graphicx}
\graphicspath{{Figure/}}
\usepackage{caption}
\usepackage{float}

\usepackage{natbib}
\usepackage{xurl}

\usepackage{algpseudocode}

\usepackage{xcolor}

\definecolor{linkblue}{RGB}{24,64,112}
\usepackage[colorlinks=true,allcolors=linkblue,breaklinks=true,pdfencoding=auto,psdextra]{hyperref}
\hypersetup{pdftitle={Sparse Equation Matching: A Derivative-Free Learning for General-Order Dynamical Systems},pdfauthor={Jiaqiang Li, Jianbin Tan, Xueqin Wang}}
\floatstyle{ruled}
\restylefloat{algorithm}
\algtext*{EndFor}
\algrenewcommand\algorithmicindent{1em}

\theoremstyle{definition}
\newtheorem{theorem}{\textbf{Theorem}}

\newtheorem{D}{Definition}
\newtheorem{Ex}{Example}

\newtheorem{remark}{Remark}

\def\bm{\boldsymbol}

\newcommand{\emailx}[1]{\href{mailto:#1}{#1}}

\title{Sparse Equation Matching: A Derivative-Free Learning for General-Order Dynamical Systems}

\author{Jiaqiang Li\\
School of Management\\
University of Science and Technology of China\\
Anhui, Hefei, China,\\
Jianbin Tan\\
Department of Biostatistics and Bioinformatics\\
Duke University, Durham, NC, USA,\\
and\\
Xueqin Wang\thanks{Email of correspondence: \emailx{wangxq20@ustc.edu.cn}}\\
School of Management\\University of Science and Technology of China\\
Anhui, Hefei, China
}

\date{}

\begin{document}

\maketitle
\begin{abstract}
Equation discovery is a fundamental learning task for uncovering the underlying dynamics of complex systems, with wide-ranging applications in areas such as brain connectivity analysis, climate modeling, gene regulation, and physical simulation. 
However, many existing approaches rely on accurate derivative estimation and are limited to first-order dynamical systems, restricting their applicability in real-world scenarios. 
In this work, we propose Sparse Equation Matching (SEM), a unified framework that encompasses several existing equation discovery methods under a common formulation. 
SEM introduces an integral-based sparse regression approach using Green’s functions, enabling derivative-free estimation of differential operators and their associated driving functions in general-order dynamical systems. 
The effectiveness of SEM is demonstrated through extensive simulations, benchmarking its performance against derivative-based approaches. 
We then apply SEM to electroencephalographic (EEG) data recorded during multiple oculomotor tasks, collected from 52 participants in a brain–computer interface experiment. 
Our method identifies active brain regions across participants and reveals task-specific connectivity patterns. 
These findings offer valuable insights into brain connectivity and the underlying neural mechanisms.
\end{abstract}
\setstretch{1.08}


\section{Introduction}

Modeling the dynamics of complex systems is central to many fields, including neuroscience \citep{Friston2009, zhang2020bayesian}, physics \citep{brunton2016discovering,ramsay2017dynamic}, epidemiology \citep{tian2021effects,tan2022transmission}, climate science \citep{brunton2016discovering, lorenz2017deterministic}, and gene regulation \citep{wu2014sparse, dai2022kernel}. 
In many of these domains, system dynamics are naturally formulated as ordinary differential equations (ODEs), which provide deterministic models for the time derivatives of dynamic processes and characterize system evolution and underlying mechanisms. 
ODEs offer a principled and flexible framework for representing continuous-time dynamics that govern many real-world phenomena. 
However, despite their widespread utility, the exact forms of the governing ODEs are often unknown or only partially understood \citep{wu2014sparse, brunton2016discovering}. 
This motivates the task of equation discovery, which aims to learn the governing equations of complex systems directly from observed data.

Formally, equation discovery for ODEs involves identifying the functional form of the governing equations:
\begin{equation}
\label{ode1}
\bm{\mathcal{P}} \bm{X}(t) = \bm{f}(\bm{X}(t), t),
\end{equation}
where \( \bm{X}(t) := (X_1(t), \dots, X_p(t))^\top \) represents the $p$-dimensional deterministic trajectories observed over time; 
\( \bm{\mathcal{P}} = (\mathcal{P}_1, \dots, \mathcal{P}_p)^\top \) is a differential operator with respect to time $t$; and 
\( \bm{f} = (f_1, \dots, f_p)^\top \) denotes an unknown deterministic driving function; the latter two components are estimated or learned from noisy observations of \( \bm{X} \).
The above formulation differs from traditional parameter estimation in dynamical systems, where the functions \( f_i \) are assumed to follow fixed parametric forms, typically informed by prior domain knowledge \citep{ramsay2007parameter, dattner2015optimal, ramsay2017dynamic, yang2021inference, tan2023age, shao2025ordinary}. 
By relaxing this assumption, equation discovery provides a more flexible, data-driven framework that can adapt to observed patterns and potentially uncover new mechanistic insights.

In the following, we present several real-world examples for the tasks of equation discovery.

\begin{Ex}[Brain Connectivity Discovery]
\label{ex:brain}
Human brain activity is often recorded using electroencephalography (EEG) data \citep{niedermeyer2005electroencephalography}, providing time-resolved signals that capture complex connectivity patterns across distinct brain regions 
\citep{Friston2011review}. 
Such connectivity can be naturally modeled using systems of ODEs, which describe how the state of each brain region evolves 
and interacts with others over time. 
A representative parametric formulation is 
\[
\frac{\mathrm{d}^K X_i(t)}{\mathrm{d} t^K}
+ b_i \frac{\mathrm{d} X_i(t)}{\mathrm{d} t}
= \sum_{j=1}^p a_{ij} X_j(t),
\qquad i = 1, \ldots, p,
\]
where \(X_i(t)\) represents the neural activity in the \(i\)th brain region, 
\(\frac{\mathrm{d}^K}{\mathrm{d} t^K}\) denotes the \(K\)th-order derivative with respect to time \(t\), 
capturing higher-order temporal dependencies in neural dynamics, and 
the coefficients \(b_i\) and \(a_{ij}\) are unknown parameters.
In this context, discovering brain connectivity corresponds to identifying the governing differential equations 
from observed EEG signals, forming a canonical equation discovery problem. 
\end{Ex}

\begin{Ex}[Climate Dynamics Discovery]
\label{ex:climate}
Climate systems are often studied using time-series data such as temperature, atmospheric pressure, and wind speed, which reflect complex, nonlinear interactions across spatial and temporal scales \citep{stein2014enso, lorenz2017deterministic}. 
These interaction mechanisms are commonly characterized by ODEs, which model how climate variables co-evolve over time and space.  
For example, climate dynamics can be modeled as
\[
\frac{\mathrm{d}^2 X(t)}{\mathrm{d}t^2} 
+ \mu \frac{\mathrm{d} X(t)}{\mathrm{d}t} 
+ \omega^2 X(t) 
= F(t),
\]
where \( X(t) \) represents temperature or atmospheric pressure, \( \mu > 0 \) denotes the damping parameter, \( \omega \) is the natural frequency, and \( F(t) \) describes external forcing.
Recently, equation discovery methods have been successfully applied to uncover ODEs in climate systems, including recovery of canonical structures such as the Lorenz system~\citep{brunton2016discovering} and the Van der Pol oscillator~\citep{owens2023data}.
\end{Ex}

\begin{Ex}[Interaction Law Discovery] 
\label{ex:inter}
Inferring interaction laws from agent-based dynamic data is a fundamental problem across many scientific disciplines~\citep{lu2019nonparametric}. 
These systems, composed of multiple interacting agents, often exhibit dynamics that depend strongly on pairwise distances between agents, 
such as particle interactions in physics \citep{carrillo2017review}. 
A representative interaction model can be expressed as
\[
\frac{\mathrm{d}^K X_i(t)}{\mathrm{d} t^K}
= \frac{1}{p}\sum_{j=1}^p 
\phi\!\big(|X_j(t) - X_i(t)|\big),
\qquad i = 1, \ldots, p,
\]
where \(X_i(t)\) denotes the position of agent \(i\), and \(\phi(\cdot)\) represents the interaction law that determines the strength and direction of influence as a function of pairwise distance.
Discovering such laws aligns closely with the broader goal of equation discovery, enabling mechanistic understanding of complex systems. 
\end{Ex}

To solve the above equation discovery tasks, a widely adopted approach is to regress the derivatives of observed trajectories onto the trajectories themselves \citep{wu2014sparse, brunton2016discovering, zhang2017bayesian, lu2019nonparametric}. 
These methods view equation discovery for ODE~\eqref{ode1} as a continuous-time analogue of autoregressive modeling \citep{shumway2000time}, where discrete temporal differences in the autoregressive framework are replaced by infinitesimal derivatives in ODEs. 
From this perspective, accurate estimation of trajectory derivatives becomes crucial, as they serve as the response variables in the equation regression framework.

Despite the usefulness of this regression-based paradigm, most existing studies focus on equation discovery for first-order systems, that is, learning driving functions associated with first-order derivatives of the data. 
Although such methods can, in principle, be extended to higher-order systems, they require estimation of higher-order derivatives to learn the corresponding driving functions. 
This requirement poses significant challenges for equation discovery, since higher-order derivative estimation is typically unstable and highly sensitive to noise \citep{wahba1990spline}.

To mitigate the reliance on derivative estimation, integration-based regression approaches have been applied in various equation discovery settings \citep{chen2017network, messenger2021weak, qian2022d, dai2022kernel}. 
These methods use integral formulations to reformulate differential equations, thereby reducing dependence on first-order derivatives in the learning task. 
However, for general-order dynamical systems, simple integration is often insufficient to fully eliminate the need for derivative estimation. 
This limitation hinders the applicability of equation discovery in settings involving complex higher-order dynamics, which commonly arise in applications such as brain connectivity identification, climate dynamics modeling, and interaction law inference \citep{zhang2020bayesian, owens2023data, lu2021learning}, as illustrated in Examples~\ref{ex:brain}--\ref{ex:inter}.

To address these challenges, we propose a novel equation discovery method tailored to general-order dynamical systems, generalizing the Green's matching approach \citep{tan2024green} from parameter estimation to differential equation learning. 
Our method employs extended integral formulations for general-order ODEs using Green's functions \citep{duffy2015green}, providing a foundation for converting ODEs into integral equations of arbitrary order. 
This transformation unifies existing derivative-based and integration-based approaches \citep{brunton2016discovering, chen2017network, brunton2022data, dai2022kernel} within a single framework, enabling a broader class of equation discovery methods that were previously unavailable. 
The unified framework provides a direct mechanism to remove derivatives from the governing equations, leading to a new equation discovery method that is entirely free of derivative estimation.

Subsequently, we extend the concept of sparsity in regression models \citep{hastie2017elements} to our framework, in line with existing work on equation discovery \citep{brunton2016discovering, brunton2022data}. 
We refer to our method as Sparse Equation Matching (SEM), an integral-based sparse regression approach for the statistical learning of differential equations. 
A key distinction of SEM from existing methods is its ability to estimate both the differential operator \(\bm{\mathcal{P}} \) of linear form and the driving function \( \bm{f} \) in~\eqref{ode1} simultaneously, without explicit estimation of data derivatives. 
This results in a more accurate and robust framework for equation discovery in general-order dynamical systems.

We apply SEM to electroencephalographic (EEG) data collected during oculomotor tasks from 52 participants in a brain--computer interface experiment \citep{cho2017eeg}. 
We first demonstrate the effectiveness of SEM in predicting EEG signals across three oculomotor tasks: eye blinking, horizontal eye movement, and vertical eye movement. 
We then use SEM to infer population-level brain networks from the EEG data, uncovering both distinct and overlapping connectivity patterns associated with each task. 
These findings provide insight into the neural mechanisms underlying oculomotor behavior and highlight the potential of SEM to advance brain connectivity analysis.

The remainder of this article is organized as follows. 
Section~\ref{Sec: method} introduces the general equation matching framework for higher-order dynamical systems. 
Building upon this foundation, Section~\ref{Sec: SEM} presents the proposed SEM method, including its algorithmic implementation and key estimation properties. 
Section~\ref{Sec: simulation} evaluates the performance of SEM through simulation studies, comparing it with derivative-based approaches. 
Section~\ref{Sec: data} applies SEM to EEG data collected during oculomotor tasks, demonstrating its capability for both signal prediction and brain connectivity inference. 
Finally, Section~\ref{Sec: discuss} concludes the paper and discusses potential directions for future research.

\section{Methodology}
\label{Sec: method}
\paragraph*{Notation} Denote \( \frac{\mathrm{d}^k}{\mathrm{d}t^k} \) for the \( k \)-th derivative of a function.
Let \( L^2(\mathcal{T}) \) be the space of square-integrable functions on a domain \( \mathcal{T} \), equipped with the standard \( L^2 \) norm \( \| \cdot \|_{L^2} \) and inner product \( \langle \cdot, \cdot \rangle \).
Define the null space of an operator \( \mathcal{P} \) as \( \operatorname{Ker}(\mathcal{P}) := \{ g(\cdot) \mid \mathcal{P}g(\cdot)\ \text{is a zero function} \} \).
Let \( \mathbb{I}(\cdot) \) denote the indicator function and \( \delta(\cdot) \) denote the Dirac delta function concentrated at zero. 
Define \( (\cdot)_+ \) as \( \max\{\cdot,0\} \).
We use the notation \( \|\cdot\| \) to denote the Euclidean norm. In what follows, we may simplify a function \( f(\cdot) \) as \( f \).

We introduce some preliminary concepts relevant to reproducing kernel Hilbert spaces (RKHS). 
Let \( \mathcal{H} \) be a Hilbert space of functions defined on a domain \( \mathcal{T} \), equipped with the inner product \( \langle \cdot, \cdot \rangle_{\mathcal{H}} \) and the corresponding norm \( \| \cdot \|_{\mathcal{H}} \). 
The space \( \mathcal{H} \) is called a RKHS if there exists a kernel function \( \mathcal{K}: \mathcal{T} \times \mathcal{T} \rightarrow \mathbb{R} \) such that \( \mathcal{K}(t, \cdot) \in \mathcal{H} \) for all \( t \in \mathcal{T} \), and the reproducing property holds:
\[
f(t) = \langle f, \mathcal{K}(t, \cdot) \rangle_{\mathcal{H}}, \quad \forall f \in \mathcal{H},\ t \in \mathcal{T}.
\]
Since \( \mathcal{K} \) is uniquely associated with the space \( \mathcal{H} \), we refer to \( \mathcal{K} \) as the reproducing kernel of the Hilbert space and denote \( \mathcal{H} \) by \( \mathcal{H}(\mathcal{K}) \).

\paragraph*{Basic Setting} Assume we collect the observed data \( \{Y_{ij}\}_{i=1,\dots,p,\ j=1,\dots,n} \), which follow:
\begin{align}
    \label{measure}
    Y_{ij} = X_i(t_j) + \epsilon_{ij},
\end{align}
for \( i = 1, \dots, p \), \( j = 1, \dots, n \), and \( t_j \in [0, C] \), where \( \epsilon_{ij} \) are mean-zero white noise terms, \( X_i(t) \) denotes the value of the trajectory \( X_i \) at time \( t \), \( \{t_j\}_{j=1}^n \subset [0,C] \) are the observed time points for \( X_i \), \( n \) is the number of observed time points, and \( p \) is the number of trajectories.  

Let \( X_i(t),\ i = 1, \dots, p \), be smooth functions over \( t \in [0, C] \) that satisfy the differential equations:
\begin{equation}
\label{ode}
\mathcal{P}^K_i X_i(t) = f_i(\bm{X}(t),t), \quad i = 1, \dots, p,
\end{equation}
where \( \bm{X}(t) := (X_1(t), \dots, X_p(t))^\top \) denotes the vector of state variables at time \( t \), and \( f_i \), \( i = 1, \dots, p \), represent the driving functions governing the dynamics of \( \bm{X} \).  
The differential operator \( \mathcal{P}_i^K \) acts on \( X_i(t) \) with respect to time \( t \) and is defined as
\begin{align}
    \label{diff-op}
    \mathcal{P}_i^K 
    = \frac{\mathrm{d}^K}{\mathrm{d}t^K} 
    + \sum_{l=1}^{K-1} \omega_{il} \frac{\mathrm{d}^l}{\mathrm{d}t^l},
\end{align}
which is a linear differential operator of order \( K \).  
Such forms of differential operators have been adopted in many real-world applications \citep{meirovitch2010fundamentals, ramsay2017dynamic, zhang2020bayesian, shao2025ordinary}.  
We assume that \( K \) is fixed and known. Here, the coefficients \( \omega_{il} \) are treated as unknown parameters.

Let \( \bm{f} := (f_1, \dots, f_p)^{\top} \), \( \bm{\omega}_i := (\omega_{i1}, \dots, \omega_{i(K-1)})^{\top} \), and \( \bm{\omega} := (\bm{\omega}_1^{\top}, \dots, \bm{\omega}_p^{\top})^{\top} \). 
Our goal is to identify the driving functions \( \bm{f} \) and the coefficients \( \bm{\omega} \) given the observed dynamic data \(Y_{ij} \)'s—a standard equation discovery task.

\subsection{Equation Matching for General-Order Systems}
\label{EM-general}

In this subsection, we propose a new estimation framework for the equation discovery task. 
Denote 
\(
\frac{\mathrm{d}^l\bm{X}}{\mathrm{d}t^l} := 
\left(\frac{\mathrm{d}^l X_1}{\mathrm{d}t^l}, \dots, \frac{\mathrm{d}^l X_p}{\mathrm{d}t^l}\right)^\top
\)
as the vector of the \( l \)-th derivatives of the dynamic trajectories.  
First, we construct an operator \( \mathcal{F}_{\bm{f}, \bm{\omega}} \) that maps the function \( \bm{X} \) and its derivatives 
\( \frac{\mathrm{d}\bm{X}}{\mathrm{d}t}, \dots, \frac{\mathrm{d}^K\bm{X}}{\mathrm{d}t^K} \)
to the zero vector:
\begin{equation}
\label{EM}
\mathcal{F}_{\bm{f},\bm{\omega}}\left(\bm{X}, \frac{\mathrm{d}\bm{X}}{\mathrm{d}t}, \dots,\frac{\mathrm{d}^K\bm{X}}{\mathrm{d}t^K},t\right) = \bm{0}, 
\quad \forall\, t \in [0, C],
\end{equation}
where $\bm{f}$ and $\bm{\omega}$ are taken as their true values.  
Since the operator \( \mathcal{F}_{\bm{f}, \bm{\omega}} \) explicitly depends on \( \bm{f} \) and \( \bm{\omega} \), it imposes constraints on the trajectory \( \bm{X} \) and its derivatives. 
Following this, we construct a loss function by integrating these equation constraints across time:
\begin{align}\label{loss_EM}
    \int_{0}^C 
    \left\| 
    \mathcal{F}_{\bm{f}, \bm{\omega}}\left( 
    \bm{X}, \frac{\mathrm{d}\bm{X}}{\mathrm{d}t}, \dots, \frac{\mathrm{d}^K\bm{X}}{\mathrm{d}t^K}, t 
    \right) 
    \right\|^2 
    \nu(t)\, \mathrm{d}t,
\end{align}
where \( \nu(\cdot) \) is a positive weighting function assigning a weight to each time point \( t \).  
Note that the loss equals zero at the true $\bm{f}$ and $\bm{\omega}$, and it is generally nonzero for other values when $\bm{f}$ and $\bm{\omega}$ if the equation are identifiable. 
Therefore, the true $\bm{f}$ and $\bm{\omega}$ can be extracted as the minimizers of the loss in~\eqref{loss_EM}, 
provided that \( \bm{X}, \frac{\mathrm{d}\bm{X}}{\mathrm{d}t}, \dots, \frac{\mathrm{d}^K\bm{X}}{\mathrm{d}t^K} \) are fully observed.

The above procedure is referred to as equation matching for ODE models, which provides a general framework for estimating ODE systems by matching their equations with the true observed trajectories. 
This framework was first proposed for parameter estimation in differential equations \citep{tan2024green}, and we here extend it to a more general equation discovery setting, where $\bm{f}$ is not assumed to follow a fixed parametric form as in \citep{tan2024green}.  
One key benefit of this framework lies in its computational efficiency. 
Specifically, the approach involves matching the trajectories 
\( \bm{X}, \frac{\mathrm{d}\bm{X}}{\mathrm{d}t}, \dots, \frac{\mathrm{d}^K\bm{X}}{\mathrm{d}t^K} \)—which are typically estimated from observed data \( Y_{ij} \)—to the constraint in \eqref{EM}. 
This enables statistical estimation without explicitly solving the ODE system, as required in conventional ODE inference methods \citep{ramsay2007parameter}, reducing the computational burden associated with ODE solving \citep{tan2024green}.

In the following, we introduce the estimation steps and provide examples for equation matching.

\paragraph*{Pre-estimation of Trajectories and Their Derivatives} 
The equation matching framework requires estimating 
\( \frac{\mathrm{d}\bm{X}}{\mathrm{d}t}, \dots, \frac{\mathrm{d}^K\bm{X}}{\mathrm{d}t^K} \) 
from the observed data \( Y_{ij} \).  
This can be accomplished using RKHS regression.  
Specifically, estimators for \( \bm{X} \), denoted as 
\( \hat{\bm{X}} = (\hat{X}_1, \dots, \hat{X}_p)^{\top} \), 
can be obtained by solving:
\begin{equation}
\label{eq:estimator_optimization}
\hat{X}_i = 
\arg\min_{f \in \mathcal{H}(\mathcal{K})} 
\left\{ 
\frac{1}{n} \sum_{j=1}^n \big( Y_{ij} - f(t_j) \big)^2 
+ \nu_i \left\| P f \right\|_{\mathcal{H}}^2 
\right\},
\end{equation}
for \( i = 1, \dots, p \),  
where \( P \) denotes a penalization operator from \( \mathcal{H} \) to \( \mathcal{H} \), 
and \( \nu_i \) is a tuning parameter that controls the strength of regularization.  
When \( \operatorname{Ker}(P) \) is a finite-dimensional space, 
the solution to \eqref{eq:estimator_optimization} with \( \nu_i > 0 \) is unique 
and can be obtained via a finite-dimensional optimization problem \citep{hsing2015theoretical}.  
The tuning parameters \( \nu_i \) can be selected using generalized cross-validation (GCV); 
see Section~3.2.2 of \citet{gu2013smoothing} for details.

One typical RKHS is the Sobolev space:
\[
\mathcal{W}_q(\mathcal{T})
:= 
\left\{
f \in L^2([0, C]) 
\ \middle| \   
\frac{\mathrm{d}^q f}{\mathrm{d}t^q} \in L^2([0, C])
\right\},
\]
where the reproducing kernel is 
\(
\mathcal{K}(t,s)
:= 
\sum_{l=0}^{q-1} \frac{t^l}{l!} \frac{s^l}{l!}
+ 
\int_0^C 
\frac{(t-u)_+^{q-1}}{(q-1)!} 
\frac{(s-u)_+^{q-1}}{(q-1)!} 
\, \mathrm{d}u,
\)
for \( q \geq 1 \), and \( \mathcal{T} = [0, C] \).  
The Sobolev space is a common functional space used for estimating latent smooth trajectories from noisy observations 
\citep{gu2013smoothing, hsing2015theoretical, tan2024functional}.  
When adopting this functional space, we require that \( q \leq K \), 
since \( \bm{X} \) is known to be \( K \)-times differentiable based on \eqref{ode}. 

\begin{remark}[Choice of the Reproducing Kernel] 
The choice of the reproducing kernel \( \mathcal{K} \) plays a key role in controlling the smoothness and flexibility of the estimated functions.  
The Sobolev kernel introduced above promotes global smoothness in the \( q \)-th derivative, consistent with the differentiable structure of the underlying dynamical equations.   
In this article, we focus on the Sobolev kernel with \( q = K \), ensuring that the smoothness order of the kernel aligns with the order of the underlying differential operator.
\end{remark}

To penalize functions in the Sobolev space, we specifically consider the penalty  
\(
\| P f \|_{\mathcal{H}}^2 
= 
\left\| \frac{\mathrm{d}^q f}{\mathrm{d}t^q} \right\|_{L^2}^2,
\)
controlling the roughness of the \( q \)-th order derivative of the curve.  
Under this penalty, the optimization problem in \eqref{eq:estimator_optimization} 
can be efficiently solved using penalized spline regression \citep{wahba1990spline}.

\begin{remark}[Asymptotic Properties of Derivative Estimators via Penalized Splines]\label{RE_splines}
The asymptotic properties of penalized splines have been extensively studied in the literature. 
Consider $X_i \in \mathcal{W}_q([0,C])$ to be $q$-times differentiable, where $k \in \{0,1,\ldots,q-1\}$ denotes the derivative order of interest. 
Suppose the observation times are equally spaced, and the errors $\varepsilon_{ij}$ are mean-zero random variables satisfying suitable moment conditions. 
Then a general result is
\[
\bigl\|\hat X_i^{(k)} - X_i^{(k)}\bigr\|_{L^2}^{2}
= O_p\!\bigl(n^{-2(q-k)/(2q+1)}\bigr),
\]
which achieves the minimax-optimal integrated $L^2$ rate for derivative estimation; see \citet{stone1980optimal, shang2013local, xiao2019asymptotic, antwi2022naive} for details.
Furthermore, \citet{zhou2000derivative} established the pointwise asymptotic normality of spline-based derivative estimators. 
For any fixed $t \in (0,C)$, they proved that
\[
\frac{\hat X^{(k)}(t) - X^{(k)}(t) - B_k(t)}
{\sqrt{\operatorname{Var}\!\bigl(\hat X^{(k)}(t)\bigr)}}
\;\overset{d}{\longrightarrow}\; N(0,1),
\]
where the bias term satisfies $B_k(t) = O\!\bigl(n^{-(q-k)/(2q+1)}\bigr)$ and
\(
\operatorname{Var}\!\bigl(\hat X^{(k)}(t)\bigr)
= O\!\bigl(n^{-2(q-k)/(2q+1)}\bigr).
\)
Hence, the pointwise mean squared error also attains the previous optimal rate. 
Note that both the asymptotic variance and $L^2$ estimation error increase with the derivative order $k$, indicating that higher-order derivatives are generally estimated with lower accuracy under observational noise and discrete-time sampling.
\end{remark}

\paragraph*{Equation Matching with Denoised Trajectories} With the pre-estimated trajectories, we obtain the empirical loss for equation matching based on \eqref{loss_EM}:
\begin{align}\label{em_EM_loss}
\int_{0}^C\left\|\mathcal{F}_{\bm{f},\bm{\omega}}\left(\bm{\hat{X}}, {\frac{\mathrm{d}\hat{\bm{{X}}}}{\mathrm{d}t}}, \cdots,{\frac{\mathrm{d}^K\hat{\bm{{X}}}}{\mathrm{d}t^K}}, t\right)\right\|^2 \nu(t) \ \mathrm{d}t.
\end{align} 
This loss is then used to estimate $\bm{f}$ and $\bm{\omega}$ from the pre-estimated trajectories.

\begin{figure}[htbp]
    \centering
    \includegraphics[width=0.8\linewidth]{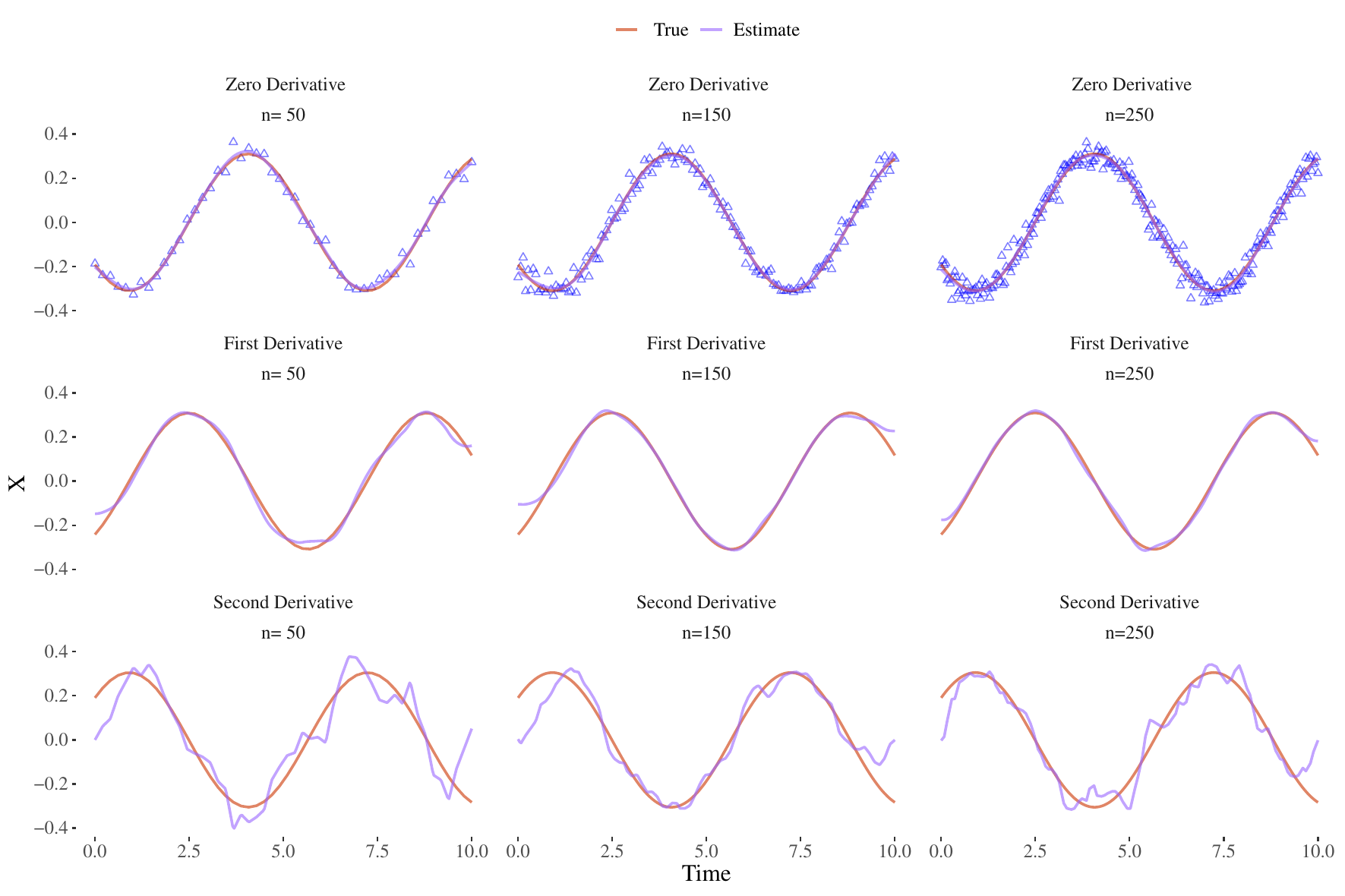}
    \caption{
    \small
    True trajectories and their derivatives, along with corresponding estimates obtained from noisy observations using RKHS regression. 
    In these examples, the signal-to-noise ratio is set to \( 0.15 \), and the sample sizes are \( n = 50, 150, 250 \), 
    with observation points evenly spaced over the time interval \([0, 10]\). 
    Blue triangles denote the observed data points.
    }
    \label{fig:denoise}
\end{figure}

A common issue with the above procedure is its potential reliance on estimated high-order derivatives, whose statistical convergence tends to deteriorate as the order \( k \) increases, as discussed in Remark~\ref{RE_splines}. 
This phenomenon is illustrated in Figure~\ref{fig:denoise} using simulated data examples, where high-order derivatives are inaccurately estimated from noisy observations. 
These inaccuracies underscore the importance of constructing \( \mathcal{F}_{\bm{f},\bm{\omega}}(\cdot) \) that minimizes dependence on high-order derivatives.

\paragraph*{Examples of Matching Operators} In this part, we relate two important equation discovery methods to our equation matching framework, each corresponding to a distinct construction of the matching operator \( \mathcal{F}_{\bm{f},\bm{\omega}}(\cdot) \).

Let \( \mathcal{F}^i_{\bm{f},\bm{\omega}}(\cdot) \) denote the \( i \)-th component of \( \mathcal{F}_{\bm{f},\bm{\omega}}(\cdot) \).  
An operator \( \mathcal{F}^i_{\bm{f},\bm{\omega}} \) directly induced by \eqref{ode} can be expressed as
\begin{align}\label{grad-match}
\mathcal{F}^i_{\bm{f},\bm{\omega}}
    &\left(\bm{X}, \frac{\mathrm{d}\bm{X}(t)}{\mathrm{d}t}, \dots, \frac{\mathrm{d}^K\bm{X}(t)}{\mathrm{d}t^K}, t \right) 
    = \frac{\mathrm{d}^K X_i(t)}{\mathrm{d}t^K} 
    + \sum_{l=1}^{K-1} \omega_{il} \frac{\mathrm{d}^l X_i(t)}{\mathrm{d}t^l} 
    - f_i(\bm{X}(t), t),
\end{align}
where equation matching reduces to a regression of the gradients
\[
\frac{\mathrm{d}^K X_i(t)}{\mathrm{d}t^K} 
+ \sum_{l=1}^{K-1} \omega_{il} \frac{\mathrm{d}^l X_i(t)}{\mathrm{d}t^l}
\]
onto \( \bm{X}(t) \) and \( t \), for all \( t \in [0, C] \).  
A well-known gradient-based method under this formulation is Sparse Identification of Nonlinear Dynamics (SINDy; \citealp{brunton2016discovering}), which applies sparse regression to recover governing equations from data by leveraging the parsimony structure of dynamical systems.  
Several extensions of SINDy have been developed to improve its accuracy and broaden its applicability \citep{kaiser2018sparse, champion2019data, fasel2022ensemble, egan2024automatically}.

An alternative approach for equation discovery is established using the Newton–Leibniz formula, which transforms the differential equation~\eqref{ode} into an integral form \citep{chen2017network, schaeffer2017sparse, dai2022kernel}.  
Specifically, integrating both sides of the ordinary differential equation yields:
\[
\frac{\mathrm{d}^{K-1}X_i}{\mathrm{d}t^{K-1}}(t)
+ \sum_{l=1}^{K-1} \omega_{il} \frac{\mathrm{d}^{l-1} X_i}{\mathrm{d}t^{l-1}}(t)
= \alpha_i + \int_0^t f_i(\bm{X}(s), s)\, \mathrm{d}s,
\]
where \( \alpha_i \) is an additional scalar parameter.  
Based on this formulation, the matching operator can be defined as
\begin{align}
   \label{integral-match}
   \mathcal{F}^i_{\bm{f},\bm{\omega},\alpha_i}
   \left(\bm{X}, \frac{\mathrm{d}\bm{X}(t)}{\mathrm{d}t}, \dots, \frac{\mathrm{d}^{K-1}\bm{X}(t)}{\mathrm{d}t^{K-1}}, t\right)= \frac{\mathrm{d}^{K-1}X_i(t)}{\mathrm{d}t^{K-1}}
   &+ \sum_{l=1}^{K-1} \omega_{il} \frac{\mathrm{d}^{l-1}X_i(t)}{\mathrm{d}t^{l-1}}
   - \alpha_i\\
   \nonumber
   &- \int_0^t f_i(\bm{X}(s), s)\, \mathrm{d}s.
\end{align}
The equation matching method defined by \eqref{integral-match} is known as integral matching \citep{ramsay2017dynamic, dattner2015optimal}.  
In contrast to the gradient-based formulation in \eqref{grad-match}, the integral matching approach introduces auxiliary parameters \( \alpha_i \), which effectively remove dependence on the highest-order derivative \( \frac{\mathrm{d}^K\bm{X}}{\mathrm{d}t^K} \).

In the special case of first-order dynamics (\( K = 1 \)), integral matching completely eliminates the need for derivative estimation.  
However, for higher-order dynamics (\( K > 1 \)), direct integration as described above does not fully remove the requirement to estimate derivatives of the underlying trajectories.  
This limitation may introduce additional inaccuracies in equation matching for general higher-order systems.

\subsection{Equation Matching with Green's Functions}

In this section, we unify gradient matching and integral matching within a common framework—termed general-order equation matching—by leveraging the Green’s function of differential operators.  
Formally, let \( \mathcal{P} \) be a differential operator defined on functions over the interval \( [0, C] \).  
The Green’s function associated with \( \mathcal{P} \) is a bivariate kernel function \( G(t, s) \) defined on \( [0, C]^2 \), which satisfies
\begin{align}
    \label{green-def}
    \mathcal{P} G(t, s) = \delta(t - s),
\end{align}
where \( \mathcal{P} \) acts on the variable \( t \).  
Given \( G(t, s) \), the solution \( X(t) \) to the equation \( \mathcal{P} X(t) = f(t) \) can be expressed as
\[
X(t) = g(t) + \int_{0}^{C} G(t, s)\, f(s)\, \mathrm{d}s,
\]
where \( g \) is any function in \( \mathrm{Ker}(\mathcal{P}) \).  
In this framework, the Green’s function serves as the inverse of the operator \( \mathcal{P} \), effectively reducing the differential order and representing the solution in an integral form.

When \( \mathcal{P} \) is the \( k \)-th order derivative operator \( \frac{\mathrm{d}^{k}}{\mathrm{d} t^{k}} \),  
the corresponding Green’s function can be explicitly written as
\begin{align}
    \label{standard-green}
    G^{k}(t, s) = \frac{(t - s)^{k - 1}}{(k - 1)!}\, \mathbb{I}(t \geq s), \quad k \geq 1.
\end{align}
This expression follows directly from the definition of the Green’s function.  
For \( k = 0 \), we define the zero-order Green’s function as \( G^{0}(t, s) = \delta(t - s) \).  
In the following, we employ these Green’s functions to reduce the order of the differential equation~\eqref{ode};  
see the next theorem.

\begin{theorem}[Integral Form of Differential Equations]\label{theo:inte}
Assume that each function \( f_i(\bm{x}, t) \) in \eqref{ode} is continuous in \( t \) and locally Lipschitz continuous in \( \bm{x} \):
\begin{itemize}
    \item For every \( \bm{x} \in \mathbb{R}^p \), the map \( t \mapsto f_i(\bm{x}, t) \) is continuous on \( [0, C] \).
    \item For every \( (\bm{x}_0, t_0) \in \mathbb{R}^p \times [0, C] \), there exist constants \( r > 0 \), \( \epsilon > 0 \), and \( L > 0 \) such that for all \( t \in [0, C] \) with \( |t - t_0| < \epsilon \) and all \( \bm{x}, \bm{y} \in \mathbb{R}^p \) satisfying \( \|\bm{x} - \bm{x}_0\| < r \) and \( \|\bm{y} - \bm{x}_0\| < r \),
    \[
    |f_i(\bm{x}, t) - f_i(\bm{y}, t)| \le L\,\|\bm{x} - \bm{y}\|.
    \]
\end{itemize}
\noindent Then, for any integer \( k \) such that \( 1 \leq k \leq K - 1 \), the differential equation \eqref{ode} is equivalent to the following integral equation:
\begin{align}  
\label{match-k}
    \frac{\mathrm{d}^{\scriptstyle K-k}X_i}{\mathrm{d}t^{ \scriptstyle K-k}}&(t) +  \sum_{\scriptstyle l=k+1}^{K-1} \omega_{il} \frac{\mathrm{d}^{\scriptstyle l-k} X_i}{\mathrm{d}t^{ \scriptstyle l-k}}(t)+ \sum_{l=1}^{ \min (k,K-1)} \omega_{il}\cdot\\
     \nonumber
  \int_0^C G^{k-l}(t, s) & X_i(s)\, \mathrm{d}s = g_{ik}(t) + \int_0^C G^k(t,s) f_i(\bm X(s),s)\, \mathrm{d}s, 
\end{align}
where \( g_{ik} \) is a function in \( \mathrm{Ker}\!\left(\frac{\mathrm{d}^k}{\mathrm{d}t^k}\right) \). Here, 
\begin{itemize}
    \item For \( k > K \), we replace the term \( \frac{\mathrm{d}^{K-k}X_i}{\mathrm{d}t^{K-k}}(t) \) in \eqref{match-k} with its integral counterpart \( \int_0^C G^{k-K}(t, s)\, X_i(s)\, \mathrm{d}s \).  
\item When \( k \geq K - 1 \), we remove the term \( \sum_{l = k+1}^{K-1} \omega_{il} \frac{\mathrm{d}^{l-k} X_i}{\mathrm{d}t^{l-k}}(t) \) from \eqref{match-k}.  
\item We omit the term \( \sum_{l = 1}^{\min(k, K-1)} \omega_{il} \int_0^C G^{k-l}(t, s)\, X_i(s)\, \mathrm{d}s \) when \( k = 0 \).
\end{itemize}    
\end{theorem}

Theorem~\ref{theo:inte} generalizes Equation~(12) in \citet{tan2024green} to integral equations of arbitrary order, which can be derived by applying the Green’s function of the differential operator \( \frac{\mathrm{d}^{k}}{\mathrm{d} t^{k}} \) to both sides of \eqref{ode}.  
When \( k = 0 \) and \( k = 1 \), this formulation reduces to the original differential equation~\eqref{ode} and its Newton–Leibniz integral form, respectively.  
By increasing \( k \) beyond \( 1 \), and even beyond the true order \( K \), one can obtain an equivalent integral equation that only involves derivatives of lower order.

Parallel to the relation between \eqref{EM} and \eqref{loss_EM}, we define the equation matching procedure based on \eqref{match-k} to estimate $\bm{f}$ and $\bm{\omega}$ as follows.

\begin{D}[Order-$k$ Equation Matching]
For \( k \geq 0 \), define the matching operator \( \mathcal{F}^i_{\bm{f},\bm{\omega},g_{ik}} \) according to \eqref{match-k} as
\begin{align}
\label{k-match}
    &\mathcal{F}^i_{\bm{f},\bm{\omega},g_{ik}}
    \left(\bm{X}, \frac{\mathrm{d}\bm{X}}{\mathrm{d}t}, \dots, \frac{\mathrm{d}^{K-k}\bm{X}}{\mathrm{d}t^{K-k}}, t\right)\\
    :=\ &\frac{\mathrm{d}^{K-k}X_i}{\mathrm{d}t^{K-k}}(t)
    + \sum_{l=k+1}^{K-1} \omega_{il} \frac{\mathrm{d}^{l-k} X_i}{\mathrm{d}t^{l-k}}(t) + \sum_{l=1}^{\min(k, K-1)} \omega_{il} \int_0^C G^{k-l}(t, s)\, X_i(s)\, \mathrm{d}s
    - g_{ik}(t)\nonumber\\
    & - \int_0^C G^k(t, s)\, f_i(\bm{X}(s), s)\, \mathrm{d}s.\nonumber
\end{align}
The corresponding loss function is defined as
\begin{align}
\label{k-match-loss}
    \int_{0}^{C} \sum_{i=1}^{p} 
    \left|
    \mathcal{F}^i_{\bm{f},\bm{\omega},g_{ik}}
    \left(\bm{X}, \frac{\mathrm{d}\bm{X}}{\mathrm{d}t}, \dots, \frac{\mathrm{d}^{K-k}\bm{X}}{\mathrm{d}t^{K-k}}, t\right)
    \right|^2 
    \nu(t)\, \mathrm{d}t,
\end{align}
where \( \nu(t) \) is a positive weighting function.
We refer to the equation matching procedure based on this loss as order-$k$ equation matching.
\end{D}

\begin{remark}[Connections to Gradient- and Integral-Based Methods for Equation Discovery]
The loss function \eqref{k-match-loss} unifies two commonly used approaches: gradient matching when \( k = 0 \) and integral matching when \( k = 1 \).  
For \( k \geq 2 \), the order-\( k \) equation matching depends only on the estimated state \( \hat{\bm{X}} \) and its lower-order derivatives 
\( \left\{ \frac{\mathrm{d}^l \hat{\bm{X}}}{\mathrm{d}t^l} \right\}_{1 \leq l \leq K - k} \), 
thereby providing an equation discovery framework that mitigates dependence on high-order derivative estimation.
\end{remark}

\begin{remark}[Derivative-Free Equation with Minimal Complexity]
When \( k \geq K \), all derivative terms are eliminated from the loss function \eqref{k-match-loss}.  
However, the dimension of the space \( \mathrm{Ker}\!\left(\frac{\mathrm{d}^k}{\mathrm{d}t^k}\right) \) associated with \( g_{ik} \) increases with \( k \), thereby introducing additional model complexity in equation discovery.  
A practical and parsimonious choice is \( k = K \), which yields a derivative-free form of equation matching while maintaining minimal model complexity.  
We focus on this derivative-free equation matching framework in the subsequent sections.
\end{remark}

\section{Sparse Equation Matching}\label{Sec: SEM}

Since no parametric form is assumed for the underlying dynamical system, we cannot directly use \eqref{k-match-loss} to estimate the driving function \( \bm{f} \) from denoised trajectories.
To address this issue, we extend the sparse equation discovery framework in \citet{brunton2016discovering} to our equation matching framework, which involves constructing a candidate set of basis functions of \( \bm{X} \) to provide a prior structure for the dynamical system.

Specifically, we define
$$
\bm H:\mathbb R^{p+1}\to\mathbb R^{D},\quad (\bm{X}^\top(t),t)^\top\mapsto\bm{H}(\bm X(t),t), \quad \forall\, t\in[0,C],
$$  
where $\bm H(\cdot)$ consists of $D$ basis functions to capture the dynamics of the differential equations.  
To simplify the notation, we denote \( (\bm{X}^\top(t), t)^\top \) as \( \bm{X}(t) \) by treating time as a state variable.
Under this notation, we assume that \( f_i \) in \eqref{ode} admits the following decomposition:
\begin{eqnarray}\label{model_force}
    f_i(\bm{X}(t)) = \bm{H}^\top(\bm{X}(t)) \bm{\beta}_i,
\end{eqnarray}
where \( \bm{\beta}_i = (\beta_{i1}, \dots, \beta_{iD})^\top \) is a \( D \)-dimensional coefficient vector that captures the contribution of each basis function.  
We suppose that \( \bm{\beta}_i \) is sparse, i.e., it contains only a small number of nonzero entries.  
This reflects the parsimonious structure of dynamical systems assumed in the literature \citep{brunton2016discovering, chen2017network, fasel2022ensemble}.

Several strategies can be considered for the construction of \( \bm H(\cdot) \).  
A common approach involves constructing candidate terms based on polynomial and trigonometric functions \citep{brunton2016discovering, zhang2020bayesian, brunton2022data}—functional forms that arise from Taylor and Fourier representations of governing physical laws. 
Specifically, \( \bm{H}(\bm{X}(t)) \) can be taken as a subset of the following candidate basis functions:
\begin{align}
\label{h_candidate}
    \left(1,\ \sin(\bm{X}),\ \cos(\bm{X}),\ \bm{X},\ \bm{X}^{2},\ \ldots,\ \bm{X}^P\right)^\top,
\end{align}
which consists of a constant term, trigonometric terms $\sin(\bm X)$ and $\cos(\bm X)$, and polynomial terms $\bm X,\bm{X}^2,\ldots,\bm {X}^P$,  
where \( \bm{X}^P \) denotes the vector of all polynomial combinations of the components of \( \bm{X} \) up to degree \( P \), that is,
\[
\bm{X}^P = \left(X_1^P, X_1^{P-1}X_2, \ldots, X_{p-1}X_p^{P-1}, X_p^P\right).
\]
Moreover, one may construct an RKHS-based model for \( f_i \) using an additive structure, where each additive component is assumed to lie in an RKHS; see \citet{dai2022kernel} for details.  
In this work, we focus on the first framework in \eqref{h_candidate} for constructing \( \bm{H}(\cdot) \).

\begin{remark}[Connections to Network Discovery of Dynamical Systems]
Model~\eqref{model_force} provides a meaningful interpretation for the task of network discovery in dynamical systems, aiming at identifying the underlying interactions among system components—often represented as a directed graph on the system variables \citep{friston2003dynamic, Friston2009, chen2017network}. 
Through the sparse estimation of $\bm{\beta}_i$ in \eqref{model_force}, we enable network discovery by expressing the dynamics through a system of equations governed by a set of basis functions. 
For example, if some of the coefficients corresponding to basis functions involving \( X_k \) are nonzero, then \( X_k \) influences the dynamics of \( X_i \). 
We then construct a directed edge from \( X_k \) to \( X_i \) to represent this influence in the system. This captures mechanisms such as regulation or information flow within the network.

\end{remark}
 
According to \eqref{k-match}, we apply the order-\( K \) equation matching equipped with the construction in \eqref{model_force}.
This yields the following matching operator: 
\begin{align}
\nonumber
   \mathcal{F}^i_{ \bm\theta_i,g_{iK}}(\bm{X}, t) :=X_i(t)+\bigg(\int_0^CG^{K-1}&(t,s)X_i(s)\mathrm{d}s,\ldots,\int_0^CG(t,s)X_i(s)\mathrm{d}s \bigg)\bm{\omega}_i\\
   \label{green-reg}
   -\bigg( \int_0^C& G^K(t, s)\, \bm H^\top(\bm{X}(s))\, \mathrm{d}s \bigg) \bm{\beta}_i- g_{iK}(t),
\end{align}
where $\bm\theta_i := (-\bm{\omega}_i^\top,\bm{\beta}_i^\top )^\top$ and $g_{iK} \in \mathrm{Ker}\left(\frac{\mathrm{d}^K}{\mathrm{d}t^K}\right)$.   
Let $\bm\Phi_K(t) = \left(1, t, \dots, t^{K-1} \right)^\top$ denote a basis for the null space $\mathrm{Ker}\!\left(\frac{\mathrm{d}^{K}}{\mathrm{d}t^{K}}\right)$. 
Then, for any $g_{iK} \in \mathrm{Ker}\!\left(\frac{\mathrm{d}^K}{\mathrm{d}t^K}\right)$, it can be expressed as 
$g_{iK}(t) = \bm{\Phi}_K^\top(t)\, \bm{\alpha}_i$ 
for some $\bm{\alpha}_i \in \mathbb{R}^K$. 
Hence, we denote $\mathcal{F}^i_{ \bm\theta_i,g_{iK}}$ by $\mathcal{F}^i_{ \bm\theta_i,\bm{\alpha}_i}$.

Define
\begin{align*}
&\bm\Theta_i(\bm  X(t)):=\bigg(\int_0^C\bm G^{K-1}(t,s)X_i(s)\mathrm{d}s,\ \int_0^C G^K(t, s)\, \bm H^\top(\bm{X}(s))\, \mathrm{d}s  \bigg),\\
    &\text{with } \int_0^C\bm G^{K-1}(t,s)X_i(s)\mathrm{d}s:=\left(\int_0^CG^{K-1}(t,s)X_i(s)\mathrm{d}s,\ldots,\int_0^CG(t,s)X_i(s)\mathrm{d}s\right).
\end{align*}
The first block $\int_0^C\bm G^{K-1}(t,s)X_i(s)\mathrm{d}s$ in $\bm\Theta_i(\bm  X(t))$ corresponds to terms from the differential operator in \eqref{ode}, while the remaining $\int_0^C G^K(t, s)\, \bm H^\top(\bm{X}(s))\, \mathrm{d}s$ corresponds to terms from the driving function $f_i$.
We then have 
\begin{align}
\label{green-match}
    \mathcal{F}^i_{\bm\theta_i,\bm\alpha_i}(\bm X, t)=X_i(t)-\bm\Theta_i(\bm X(t))\bm\theta_i-\bm\Phi_K^\top(t)\bm\alpha_i.
\end{align}
Similar to $\bm{\beta}_i$, we assume that \( \bm{\theta}_i \) is a sparse vector for each \( i \). 
This assumption enables parsimonious equation discovery for both the linear differential operator and the driving function in \eqref{ode}, 
providing flexibility in uncovering the governing equations of general-order systems.

To facilitate the sparse estimation, we estimate \( \bm{\theta}_i \) and \( \bm{\alpha}_i \) by minimizing a regularized equation matching loss:
\begin{align}
\nonumber
(\hat{\bm{\theta}}_i, \hat{\bm{\alpha}}_i) &=\arg\min_{\scriptscriptstyle \bm\theta_i\in \mathbb{R}^{K-1+D},\bm\alpha_i\in \mathbb{R}^{K}}\mathcal{L}_i(\bm\theta_i ,\bm\alpha_i \mid \bm{X})\\
    \label{loss_fcn}
&:=\arg\min_{\scriptscriptstyle \bm\theta_i\in \mathbb{R}^{K-1+D},\bm\alpha_i\in \mathbb{R}^{K}} \int_0^C \left| \mathcal{F}^i_{\bm\theta_i,\bm\alpha_i}( \bm{X}, t) \right|^2 \mathrm{d}t + R(\bm\theta_i),
\end{align}
where \( R(\bm\theta_i) \) is a sparsity-inducing penalty applied to \( \bm{\theta}_i \), $i= 1,\ldots,p.$
We approximate the integral in \eqref{loss_fcn} using dense discrete time grids. 
The minimization of \eqref{loss_fcn} then becomes a standard sparse least squares regression problem.

To enforce sparsity, one may apply the least absolute shrinkage and selection operator (LASSO) penalty \citep{hastie2017elements}, 
i.e., \( R(\bm\theta_i) = \lambda_i \|\bm\theta_i\|_1 \), where \( \|\bm\theta_i\|_1 \) denotes the \( \ell_1 \)-norm of \( \bm\theta_i \) 
and $\lambda_i$ is a tuning parameter. 
This penalty has been widely adopted in the context of equation discovery \citep{champion2019data, brunton2022data, egan2024automatically}. 
Other sparse regression techniques, such as the elastic net \citep{zou2005regularization}, group LASSO \citep{yuan2006model}, 
and adaptive best-subset selection (ABESS) \citep{zhu2020polynomial}, can also be employed for equation matching.

We refer to the regularized formulation in \eqref{loss_fcn} as \textbf{S}parse \textbf{E}quation \textbf{M}atching (SEM). 
The complete SEM procedure is outlined in Algorithm~\ref{alg_discovery}.

\begin{algorithm}[ht]
\caption{Sparse Equation Matching for General-Order Dynamical Systems} 
\label{alg_discovery} 
\begin{algorithmic}[1]
\State \textbf{Input}: Observed time points \( t_j \), \( j = 1, \dots, n \); noisy observations \( Y_{i}(t_j) \), \( i = 1, \dots, p \); 
order \( q \) of the Sobolev space; \( D \) fixed candidate basis functions \( \bm H(\cdot) \); sparsity-inducing penalty \( R(\cdot) \).
\For{\( i = 1, \dots, p \)}
    \State Estimate the trajectory \( \hat{X}_i(t) \) by solving:
    \[
    \hat{X}_i = \arg\min_{f \in \mathcal{W}_q(\mathcal T)} 
    \left\{ 
        \frac{1}{n} \sum_{j=1}^n \left( Y_{ij} - f(t_j) \right)^2 
        + \nu_i \left\| \frac{\mathrm{d}^q f}{\mathrm{d} t^q} \right\|_{L^2}^2 
    \right\},
    \]
    where \( \nu_i \) is chosen via GCV.
\EndFor
\State Construct the full trajectory \( \hat{\bm{X}}(t) = (\hat{X}_1(t), \dots, \hat{X}_p(t))^\top \).
\For{\( i = 1, \dots, p \)}
    \State For each \( t \in [0, C] \), calculate \( \bm\Phi_K(t) = (1, t, \dots, t^{K-1})^\top \) and
    \[
    \begin{aligned}
        \bm\Theta_i(\hat{\bm X}(t)) 
        :&= \bigg(
            \int_0^C G^{K-1}(t, s)\, \hat{X}_i(s)\, \mathrm{d}s,\;
            \ldots,\;
            \int_0^C G^K(t, s)\, \bm H^\top(\hat{\bm X}(s))\, \mathrm{d}s
        \bigg).
    \end{aligned}
    \]
    \State Estimate coefficients by solving the regularized equation-matching problem:
     \[
    \begin{aligned}
        & (\hat{\bm{\theta}}_i, \hat{\bm{\alpha}}_i) = \arg\min_{\bm\theta_i, \bm\alpha_i} \int_0^C\left| \hat{X}_i(t) - \bm\Theta_i(\hat{\bm X}(t)) \bm\theta_i - \bm\Phi_K^\top(t) \bm\alpha_i\right |^2 \mathrm{d}t + R(\bm\theta_i) .
    \end{aligned}
    \]
\EndFor
\State \textbf{Output}: \( \hat{\bm\theta}_i \), \( \hat{\bm\alpha}_i \) for \( i = 1, \dots, p \).
\end{algorithmic}
\end{algorithm}

For SEM with the LASSO penalty \( R(\bm\theta_i) = \lambda_i \|\bm\theta_i\|_1 \), 
the tuning parameter \( \lambda_i \) is selected via 10-fold cross-validation (CV) for each \( i \). 
Specifically, we partition the time interval \( [0, C] \) evenly into ten disjoint subintervals \( \{I_l\}_{l=1}^{10} \). 
Let \( \hat{\bm X} \) denote the estimated trajectories obtained from \eqref{eq:estimator_optimization}. 
Given a fixed value of \( \lambda_i \), the $l$-fold-specific estimates 
\( (\hat{\bm\theta}_i^{(l)}, \hat{\bm\alpha}_i^{(l)}) \) are obtained by solving:
\begin{align*}
   & \left( \hat{\bm\theta}_i^{(l)}, \hat{\bm \alpha}_i^{(l)}\right)= \arg\min_{\bm\theta_i,\bm\alpha_i} 
    \left\{ 
        \int_0^C 
        \left| 
            \mathcal{F}^i_{\bm\theta_i,\bm\alpha_i}( \hat{\bm{X}}, t) 
        \right|^2 
        \mathbb{I}(t \notin I_l)\, \mathrm{d}t 
        + \lambda_i\|\bm\theta_i\|_1 
    \right\}.
\end{align*}
The cross-validation error for \( \lambda_i \) is then computed as
\[
\mathrm{CV}(\lambda_i) 
:= 
\frac{1}{10} 
\sum_{l=1}^{10} 
\left( 
    \int_0^C 
    \left| 
        \mathcal{F}^i_{\hat{\bm\theta}_i^{(l)},\hat{\bm\alpha}_i^{(l)}}( \hat{\bm{X}}, t) 
    \right|^2 
    \mathbb{I}(t \in I_l)\, 
    \mathrm{d}t
\right).
\]
The optimal tuning parameter \( \lambda_i \) is then selected by minimizing the cross-validation error.
 
\paragraph*{Invariance of Sparse Equation Matching} Note that any function of the form
\[
\tilde{G}^k(t,s) = G^k(t,s) + \bm{\Phi}_k^\top(t)\bm{v}_k(s)
\]
is also a valid Green's function for the differential operator 
\( \frac{\mathrm{d}^k}{\mathrm{d}t^k} \),
where \( \bm{\Phi}_k(t) := (1, t, \ldots, t^{k-1})^\top \) 
denotes a basis for 
\( \mathrm{Ker}\left( \frac{\mathrm{d}^k}{\mathrm{d}t^k} \right) \), 
and \( \bm{v}_k(s) := (\psi_{k1}(s), \ldots, \psi_{kk}(s))^\top \) 
is an arbitrary function mapping \( [0, C] \) to \( \mathbb{R}^k \).  
The following result shows that the estimation of \( \bm{\theta}_i \) 
is invariant to the choice of Green's function.

\begin{theorem}[Invariance of SEM Estimation to Green's Function Choice]
\label{thm-optimizer}
Let \( \tilde{G}^k \) be any valid Green's function for the differential operator \( \frac{\mathrm{d}^k}{\mathrm{d}t^k} \).  
Given any regularization function \( R(\cdot) \) and a set of basis functions \( \bm{H}(\cdot) \), define the equation-matching operator associated with \( \{ \tilde{G}^k \}_{k=1}^K \) as:
\[
\tilde{\mathcal{F}}^i_{\bm\theta_i, \bm\alpha_i}(\bm{X},t) = X_i(t) - \tilde{\bm{\Theta}}_i(\bm{X}(t)) \bm\theta_i - \bm{\Phi}_K^\top(t)\bm\alpha_i,
\]
where \( \tilde{\bm{\Theta}}_i(\bm{X}(t)) \) is given by
\[
\begin{aligned}
\tilde{\bm{\Theta}}_i&(\bm{X}(t)) := \left(
\int_0^C \tilde{G}^{K-1}(t, s) X_i(s) \, \mathrm{d}s,\;
\ldots,\;\right.
\left.\int_0^C \tilde{G}(t, s) X_i(s) \, \mathrm{d}s,\;
\int_0^C \tilde{G}^K(t, s) \bm{H}^\top(\bm{X}(s)) \, \mathrm{d}s
\right).
\end{aligned}
\]
The corresponding SEM estimator is defined as
\[
(\tilde{\bm{\theta}}_i, \tilde{\bm{\alpha}}_i) := \arg\min_{\scriptscriptstyle \bm\theta_i \in \mathbb{R}^{K-1+D}, \bm\alpha_i \in \mathbb{R}^{K}} \int_0^C \left| \tilde{\mathcal{F}}^i_{\bm\theta_i, \bm\alpha_i}(\bm{X},t) \right|^2 \mathrm{d}t + R(\bm\theta_i).
\]
Let \( \tilde{\Theta} \) and \( \tilde{A} \) denote the sets of minimizers 
\( \tilde{\bm{\theta}}_i \) and \( \tilde{\bm{\alpha}}_i \), respectively. 
Similarly, let \( \hat{\Theta} \) and \( \hat{A} \) denote the corresponding minimizers 
from \eqref{loss_fcn} using the canonical Green's functions. 
Then there exists an identity map \( \operatorname{id} \) and a bijection \( \mathcal{M} \) such that
\[
\operatorname{id}(\tilde{\Theta}) = \hat{\Theta}, 
\quad \text{and} \quad
\mathcal{M}(\tilde{A}) = \hat{A}.
\]
\end{theorem}

Theorem~\ref{thm-optimizer} indicates that different choices of Green's functions 
only affect the null-space component \( g_{iK} \) in \eqref{green-reg}, 
but do not alter the SEM estimates \( \hat{\bm{\theta}}_i \).

\section{Simulation}
\label{Sec: simulation}
In this section, we assess the numerical performance of SEM with the LASSO penalty. 
We compare our method with a gradient-based approach using the same LASSO regularization—SINDy \citep{brunton2016discovering}—which corresponds to the order-0 sparse equation matching framework.

We compare the two methods using two examples of second-order ODEs. 
For each example, we first numerically simulate the trajectories \( X_i \), \( i = 1, \dots, p \), on an evenly spaced time grid \( \{t_j : j = 1, \dots, n\} \) over the interval \( [0, C] \). 
We then generate observations \( Y_i(t_j) \) from a normal distribution with mean \( X_i(t_j) \) and variance \( \sigma_i^2 \), where \( \sigma_i^2 = \gamma^2 \int_0^C X_i^2(t)\, \mathrm{d}t / C \), and \( \gamma \) controls the noise level. 
For comparison, both methods use the same pre-estimation procedure based on RKHS regression~\eqref{eq:estimator_optimization}, 
with \( \mathcal{H}(\mathcal{K}) = \mathcal{W}_2(\mathcal{T}) \) and penalty term 
\( \|P f\|_{\mathcal{H}}^2 = \big\| \tfrac{\mathrm{d}^q f}{\mathrm{d}t^q} \big\|_{L^2}^2 \); 
the tuning parameter is selected via GCV.  

In all cases, we consider sample sizes \( n = 50, 150, 250, 350 \), evenly spaced in \( [0, C] \), and noise levels \( \gamma = 0.05, 0.07, 0.09 \). 
Each experiment is repeated 2000 times.

In the following, we demonstrate each example of ODEs separately.

\paragraph*{Nonlinear Pendulum}
The nonlinear pendulum equation describes the motion of a single pendulum swinging under the influence of gravity, and serves as a fundamental example of a nonlinear differential equation in classical mechanics \citep{hairergeometric, thompson2002nonlinear}. 
The equation is given by:
\[
\frac{\mathrm{d}^2X(t)}{\mathrm{d}t^2} = -\sin(X(t)),
\]
where \( X(t) \) denotes the angular displacement of the pendulum at time \( t \).  
We numerically solve the nonlinear pendulum system on the interval \([0, 20]\), 
with initial conditions \((X(0), \tfrac{\mathrm{d}X(0)}{\mathrm{d}t})\) sampled from the uniform distribution \(\operatorname{U}([-0.5, 0.5]^2)\). 
In this example, we set \(\bm{H}(X) = \left(1, X, X^2, X^3, X^4, \sin(X), \cos(X)\right)^\top\) for equation discovery.

We evaluate the performance of equation discovery using the relative error rate (RER), defined as
\[
\operatorname{RER} := \frac{1}{S} \sum_{s=1}^S \frac{\|\hat{f}^{s} - f\|_2}{\|f\|_2},
\]
where \( f(\cdot) \) is the true driving function of the ODE, \( \hat{f}^{s}(\cdot) \) is the estimated driving function in the \( s \)-th replicated experiment, and \( S \) is the number of replicated simulations.  

\begin{figure}[htbp]
    \centering
    \includegraphics[width=.98\linewidth]{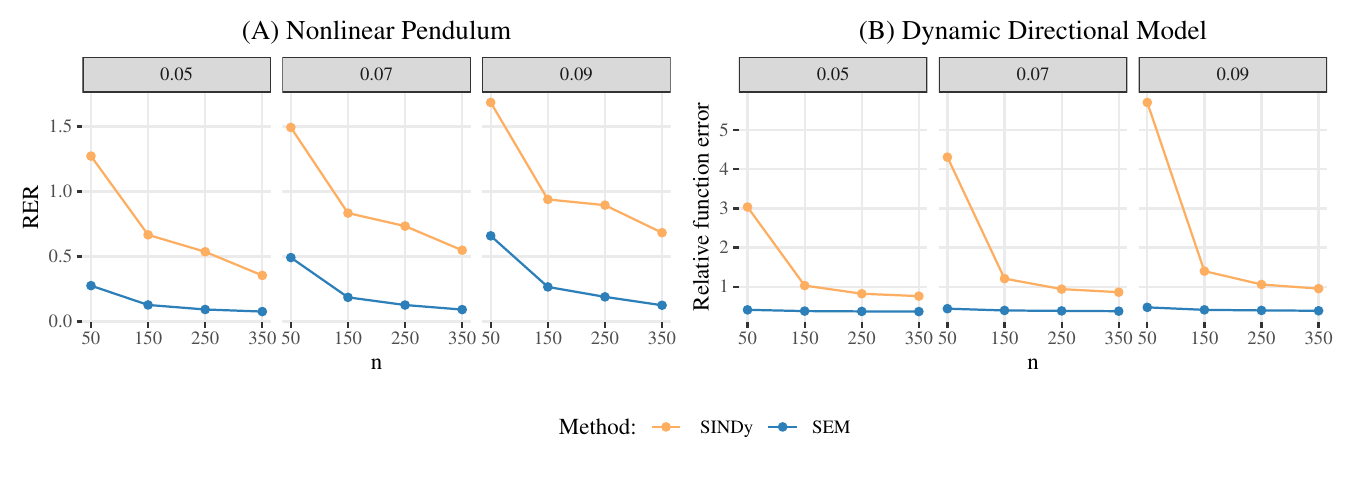}
    \caption{\small Relative error rate (RER) of the estimated driving functions under various sample sizes \( n \) and contamination levels \( \gamma \) (main title of each subgraph).}
    \label{fig:rer}
\end{figure}

The RERs of the nonlinear pendulum with \( S = 2000 \) and various values of \( n \) and \( \gamma \) are presented in Figure~\ref{fig:rer}(A). 
We observe that SEM consistently outperforms SINDy across all tested sample sizes and noise levels. 
The suboptimal performance of SINDy likely arises from inaccuracies in derivative estimation, which introduce substantial errors into the equation discovery process and reduce the fidelity of the recovered models.

To illustrate the behavior of equation discovery, we plot the vector fields corresponding to both the true and discovered dynamics of the nonlinear pendulum. 
The vector field \( \mathcal{F} \) is defined by the mapping:
\[
\mathcal{F}:\quad
\left(X(t),\frac{\mathrm{d}X(t)}{\mathrm{d}t}\right) 
\mapsto 
\left(\frac{\mathrm{d}X(t)}{\mathrm{d}t},\, f\!\left(X(t),\frac{\mathrm{d}X(t)}{\mathrm{d}t}\right)\right),
\]
where \(
\left( X(t), \tfrac{\mathrm{d}X(t)}{\mathrm{d}t} \right)
\) represents an arbitrary input of location and velocity of the trajectory, 
\( f \) denotes a function of \( X(t) \) and \( \tfrac{\mathrm{d}X(t)}{\mathrm{d}t} \) that models the second-order derivative (i.e., acceleration), and 
\(
\left( \tfrac{\mathrm{d}X(t)}{\mathrm{d}t},\, f\!\left(X(t), \tfrac{\mathrm{d}X(t)}{\mathrm{d}t} \right) \right)
\)
is the corresponding output—the velocity and acceleration of the system.  
The function \( f \) is determined either by the true differential equation (e.g., \( f\big(X(t), \tfrac{\mathrm{d}X(t)}{\mathrm{d}t}\big) = -\sin(X(t)) \)) or by an equation discovered using different methods.  
This mapping generates a field of arrows across the plane \( \mathbb{R}^2 \), indicating both the magnitude and direction of motion, thereby visually encoding the system dynamics governed by the underlying differential equations.

\begin{figure}[htbp]
    \centering
    \includegraphics[width=.95\linewidth]{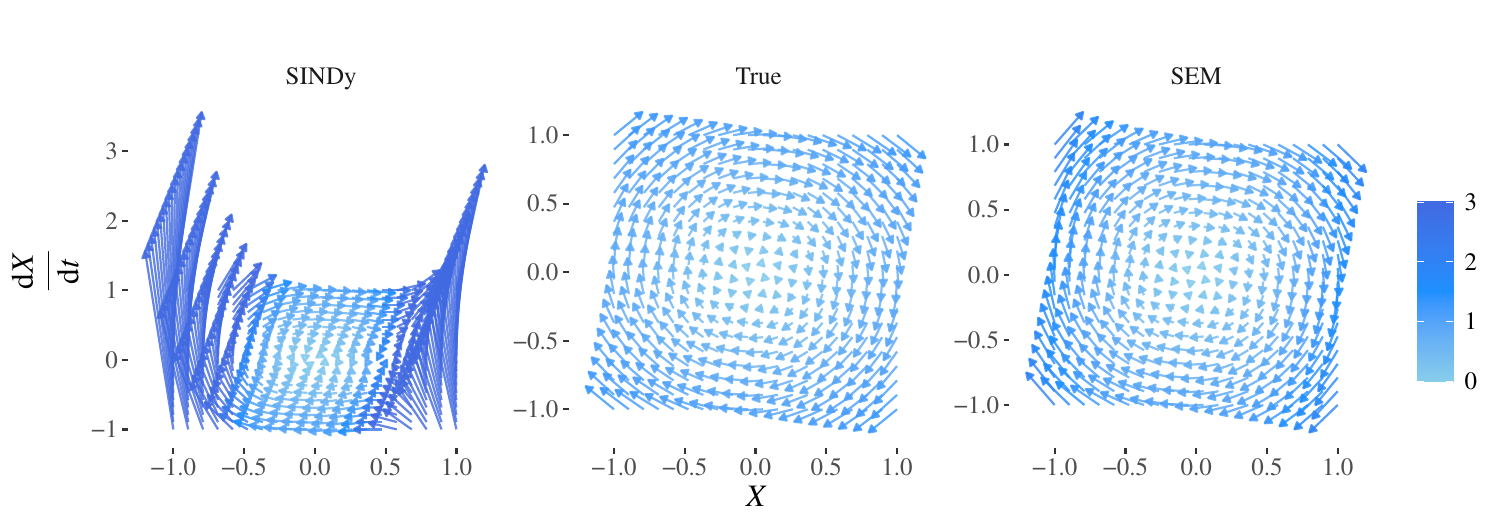}
    \caption{\small True and average estimated vector fields for the nonlinear pendulum under the setting of \( n = 150 \) and \( \gamma = 0.05 \). The color indicates the magnitude of the arrows.}
    \label{fig:vf_pen}
\end{figure}

Figure~\ref{fig:vf_pen} presents both the true vector field and the averaged estimated vector fields for the nonlinear pendulum. 
These results are based on 2000 simulation runs under the setting \( n = 150 \) and \( \gamma = 0.05 \). 
We observe that the vector field estimated using SEM closely approximates the true vector field, 
whereas the one obtained via SINDy exhibits noticeable biases and distortions in the directional flow.

In Appendix~\ref{sec_ap:NP}, we further examine the case where the candidate basis functions do not include the true components of the underlying driving function. This analysis illustrates the certain robustness of SEM when the model is misspecified or when the true functional forms are not perfectly represented in the candidate library.

\begin{figure}[htbp]
    \centering
    \includegraphics[width=.7\linewidth]{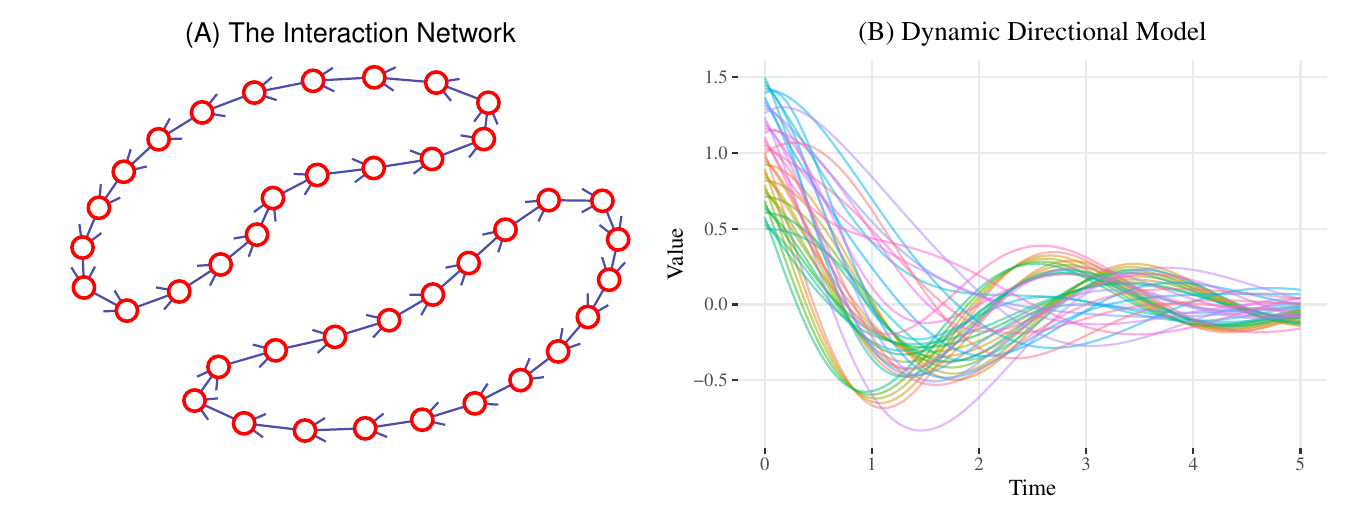}
    \caption{\small (A) The directed interaction network;  Each node represents a neural mass unit whose dynamics are governed by local excitation, inter-node directional coupling, and proportional damping that stabilizes oscillatory amplitudes. The direction between two nodes indicates that the trajectory of one node is affected by that of the other according to the dynamic directional model. (B) The trajectories generated from the dynamic directional model.}
    \label{fig:ddm_net}
\end{figure}

\paragraph*{Second-order Dynamic Directional Model}
Second-order dynamic directional models form a fundamental framework for characterizing oscillatory dynamics in neuroscience \citep{sporns2016networks, zhang2020bayesian}. We consider the following dynamic directional model:
\[
\begin{aligned}
    \frac{\mathrm{d}^2 X_i(t)}{\mathrm{d}t^2} = a_{1,i} X_i(t) + a_{1,\tau_1(i)} X_{\tau_1(i)}(t) - b_{1,i} \frac{\mathrm{d}X_i(t)}{\mathrm{d}t}&,
    \quad i = 1, \dots, &20, \\
    \frac{\mathrm{d}^2 X_i(t)}{\mathrm{d}t^2} = a_{2,i} X_i(t) + a_{2,\tau_2(i)} X_{\tau_2(i)}(t) - b_{2,i} \frac{\mathrm{d}X_i(t)}{\mathrm{d}t}&,
    \quad i = 21, \dots, &40,
\end{aligned}
\]
where \( \tau_1 \) is a permutation on \( \{1, 2, \dots, 20\} \) and \( \tau_2 \) is a permutation on \( \{21, 22, \dots, 40\} \). 
By this construction, each \( X_i \) represents a node in a directed network organized into two clusters, with nodes connected through directed loops within each cluster. 
An illustration of this directed network structure is provided in Figure~\ref{fig:ddm_net}, Panel (A).

We generate the trajectories of the above dynamic system on the interval \( [0, 5] \), with the following parameters:
\[
\begin{aligned}
    &a_{1,i} = -4, \; a_{1,\tau_1(i)} = (-1)^i \times 1.2, \;b_{1,i} = -1.3,\;\; i = 1, \dots, 20, \\
    &a_{2,i} = -3.5, \; a_{2,\tau_2(i)} = (-1)^i \times 2, \; b_{2,i} = -2,\quad i = 21, \dots, 40.
\end{aligned}
\]
We set the initial conditions as:
\[
\begin{aligned}
    X_i(0) = 1 - \frac{i-1}{38}, \quad \frac{\mathrm{d}X_i}{\mathrm{d}t}(0) = -1.5 + (-1)^i \times 0.5, &
    i = 1, \dots, &20, \\
    X_i(0) = 1.5 - \frac{i-21}{38}, \quad \frac{\mathrm{d}X_i}{\mathrm{d}t}(0) = -1.5 + \frac{2(i-21)}{19}, &
    i = 21, \dots, &40.
\end{aligned}
\]
The generated trajectories are illustrated in Figure~\ref{fig:ddm_net} (B). 
We set \( \bm{H}(\bm{X}) = \left(1, X_1, X_2, \dots, X_{40}\right)^\top \) for equation discovery in this case.

We evaluate the performance of equation discovery using the RER, similar to that defined in the nonlinear pendulum example. The RER is given by:
\[
\operatorname{RER} := \frac{1}{p}\sum_{i=1}^p\frac{1}{S} \sum_{s=1}^S \frac{\|\hat{f}_i^{s} - f_i\|_2}{\|f_i\|_2},
\]
where \( f_i(\cdot) \) is the true driving function for \( X_i \), \( \hat{f}_i^{s}(\cdot) \) is the estimated driving function in the \( s \)-th replicated experiment, and \( S = 2000 \) is the number of replications. The results for the second-order directional dynamic model under varying values of \( n \) and \( \gamma \) are presented in Figure~\ref{fig:rer}(B). 
These results align with the findings from the nonlinear pendulum example, further confirming the advantages of SEM across different types of dynamical systems.

\begin{figure}[htbp]
    \centering
    \includegraphics[width=0.95\linewidth]{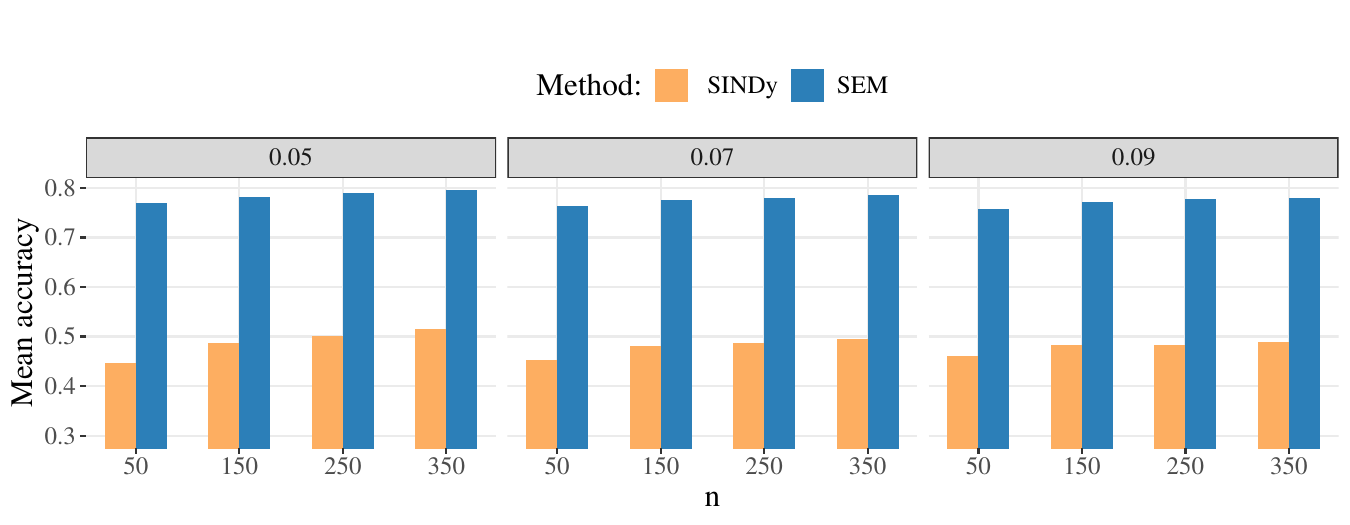}
    \caption{\small Mean accuracy of the estimated networks under different sample sizes \( n \) and noise levels \( \gamma \) (indicated in the title of each subfigure).}
    \label{fig:ddm_acc}
\end{figure}

In the dynamic directional model, we are also interested in identifying the latent directed network, as illustrated in Figure~\ref{fig:ddm_net}(A). 
Specifically, let \( E = \{e_{i,j}\}_{i,j=1}^{40} \) denote the adjacency matrix of the true directed network, where \( e_{i,j} = 1 \) if \( X_j \) contributes to the change in the second-order derivative of \( X_i \), and \( e_{i,j} = 0 \) otherwise. 
To assess estimation performance, we define the estimated adjacency matrix \( \hat{E}^s = \{\hat{e}^s_{i,j}\}_{i,j=1}^{40} \) based on the discovered equations in the \( s \)-th experiment. 
Accordingly, the mean accuracy (MA) is given by:
\[
\operatorname{MA} := \frac{1}{S} \sum_{s=1}^S \frac{\sum_{i,j=1}^{40} \mathbb{I}(\hat{e}^s_{i,j} = e_{i,j})}{40^2}.
\]
Figure~\ref{fig:ddm_acc} presents the estimation accuracy of the adjacency matrices for the second-order dynamic directional model obtained using the two methods. 
The results show that SEM recovers the network structure with substantially higher accuracy than SINDy.
 
To illustrate the behavior of equation discovery, we similarly define the vector field for the dynamic directional model. 
Taking \( X_{19} \) as an example, its vector field is defined as  
\[
\begin{aligned}
   & \mathcal{F}_{X_{19}}: \quad \mathbb{R}^2 \to \mathbb{R}^2,  \left(X_{19}(t), \frac{\mathrm{d}X_{19}(t)}{\mathrm{d}t}\right)\mapsto 
    \left(
        \frac{\mathrm{d}X_{19}(t)}{\mathrm{d}t},
        f_{19}\left(X_{1}(t), \ldots, X_{40}(t), \frac{\mathrm{d}X_{19}(t)}{\mathrm{d}t}\right)
    \right),
\end{aligned}
\]
where \( f_{19} \) denotes the driving function of \( X_{1}(t), \ldots, X_{40}(t) \) and \( \frac{\mathrm{d}X_{19}(t)}{\mathrm{d}t} \), modeling the second-order derivative (acceleration). 
This function is determined either by the true dynamical equation or by an equation discovered using different methods. 
To compute \( f_{19} \), we set \( X_{i}(t) = b \) for \( i = 1, \dots, 40 \) with \( i \neq 19 \), reflecting the acceleration of \( X_{19} \) conditional on the other trajectories taking a constant value \( b \).

\begin{figure}[htbp]
    \centering
    \includegraphics[width=0.9\linewidth]{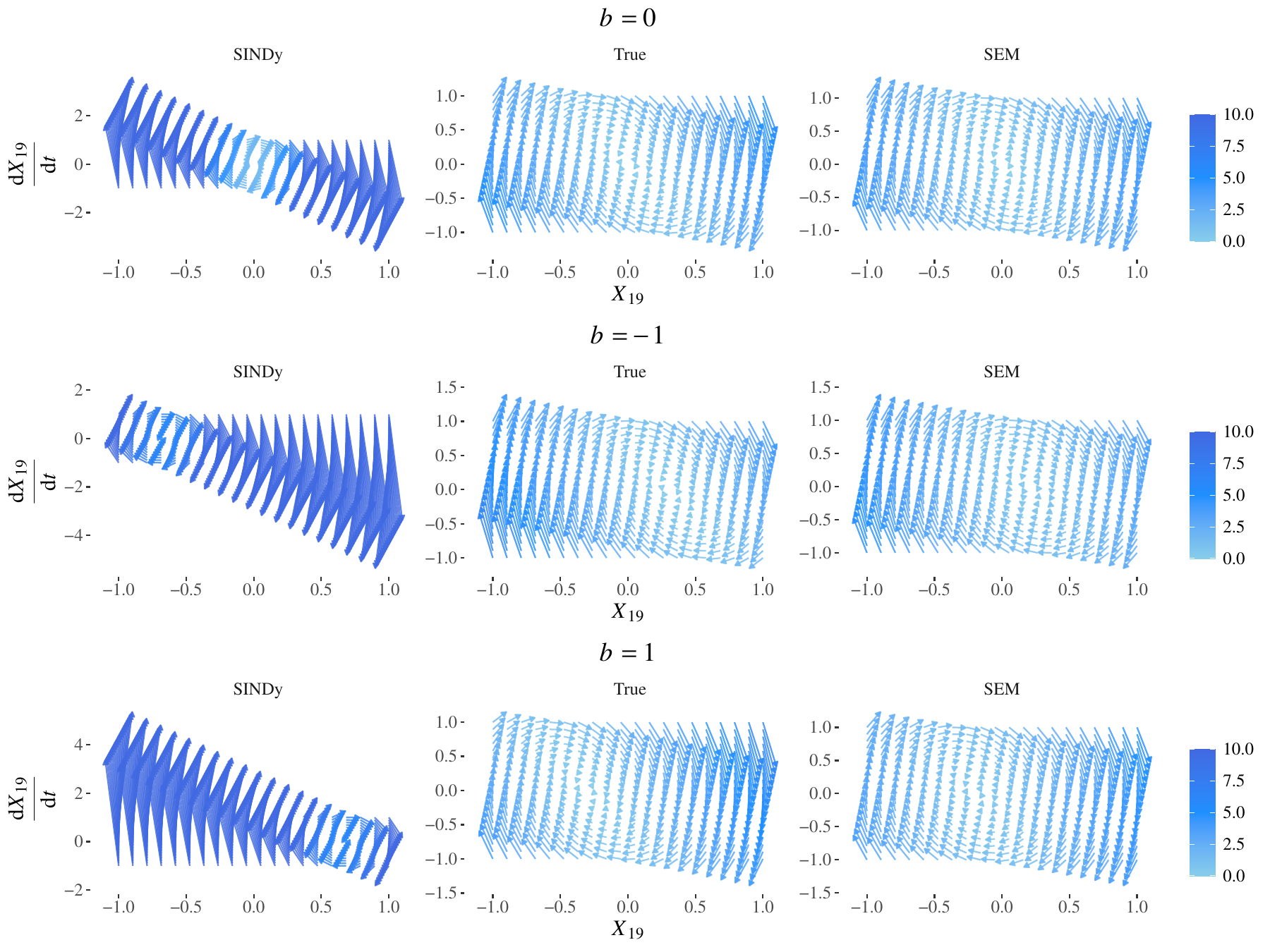}
    \caption{\small True and average estimated vector fields for \( X_{19} \) under three different values of \( b \), based on 2000 simulation runs with \( n = 150 \) and \( \gamma = 0.05 \). The color indicates the magnitude of the arrows.}
    \label{fig:vf_ddm}
\end{figure}

Figure~\ref{fig:vf_ddm} presents both the true vector fields and the mean estimated vector fields for \( X_{19} \) under three different values of \( b \), based on 2000 simulation runs with \( n = 150 \) and \( \gamma = 0.05 \). 
The results demonstrate that the vector fields estimated using SEM more closely approximate the true fields than those obtained using SINDy.

For the dynamic directional model, we need to pre-smooth multiple trajectories separately. 
To evaluate the impact of the pre-estimation step, we further compare the RKHS-based pre-estimation with a functional principal component analysis (FPCA) approach \citep{R-fdapace}, which pools all trajectories together for denoising rather than smoothing each trajectory individually. 
The FPCA procedure relies on a key assumption that \( X_1, \ldots, X_p \) are independent and identically distributed (i.i.d.) random functions, allowing the pre-smoothing step to borrow strength across curves through the shared structure.

\begin{figure}[htbp]
    \centering
    \includegraphics[width=0.55\linewidth]{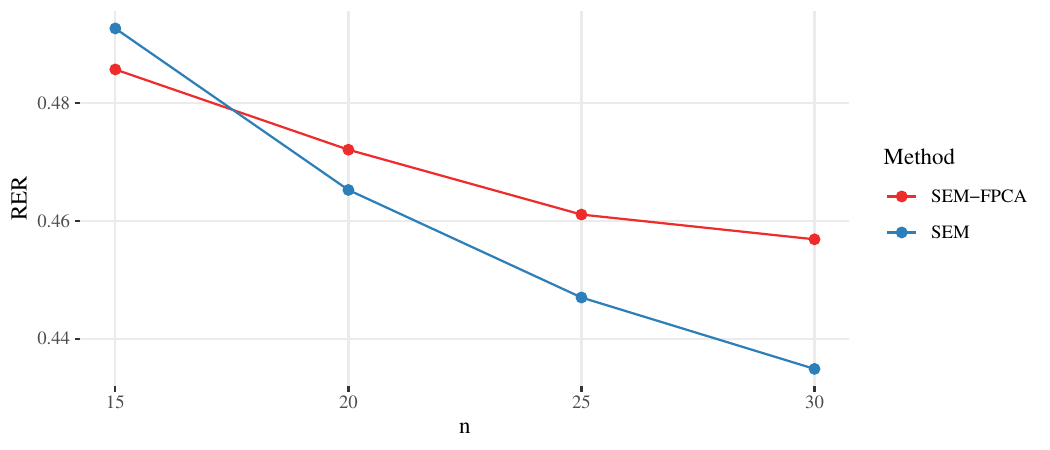}
    \caption{\small 
    Relative error rate (RER) of the estimated driving functions 
    for different numbers of observed time points \( n = 15, 20, 25, 30 \) 
    under contamination level \( \gamma = 0.05 \). 
    SEM performs pre-estimation using an RKHS approach, 
    whereas SEM-FPCA employs FPCA for pre-estimation.
    }
    \label{fig:ddm_sparse}
\end{figure}

Figure~\ref{fig:ddm_sparse} presents the relative error rates (RERs) of SEM for the estimated driving functions, obtained using the two pre-estimation methods: RKHS regression and FPCA.
When the numbers of observed time points is small (\( n = 15 \)), FPCA yields slightly better performance than RKHS regression due to the benefit of borrowing strength across trajectories. 
However, as $n$ increases, SEM with RKHS pre-estimation achieves superior performance, demonstrating the advantage of the RKHS method for densely observed trajectory data. 
Here, the inferior performance of FPCA arises because the trajectories in the dynamical system are generally not i.i.d., violating the key assumption required for FPCA.

\section{Data Analysis}
\label{Sec: data}
To demonstrate the effectiveness of SEM, we analyze electroencephalogram (EEG) data from a brain–computer interface (BCI) experiment \citep{cho2017eeg}. 
This dataset is publicly available on the GigaScience Database (\url{http://gigadb.org/dataset/100295}) and consists of 64-channel EEG recordings from 52 participants, sampled at 512~Hz (approximately every 1.95~milliseconds). 
Each participant performed three oculomotor tasks—eye blinking, horizontal eye movement, and vertical eye movement—designed to facilitate motor imagery-based BCI research. 
Additionally, resting-state EEG recordings were collected as control baselines.

Each oculomotor task spans a 5-second trial, yielding \( n = 2560 \) (\(512 \times 5\)) observations per channel. 
In contrast, the resting-state recording lasts 60 seconds, resulting in \( n = 30{,}720 \) (\(512 \times 60\)) time points per channel. 
For each task, the EEG data of the \( h \)th participant (\( h = 1, \dots, 52 \)) are denoted by \( Y_{ij}^{(h)} \), 
where \( j \in \{1, \ldots, n\} \) indexes the observation time points and \( i \in \{1, \ldots, p\} \) indexes the EEG channels, 
with \( p = 64 \) representing the total number of channels.

\begin{figure}[htbp]
   \centering
   \includegraphics[width=0.95\linewidth]{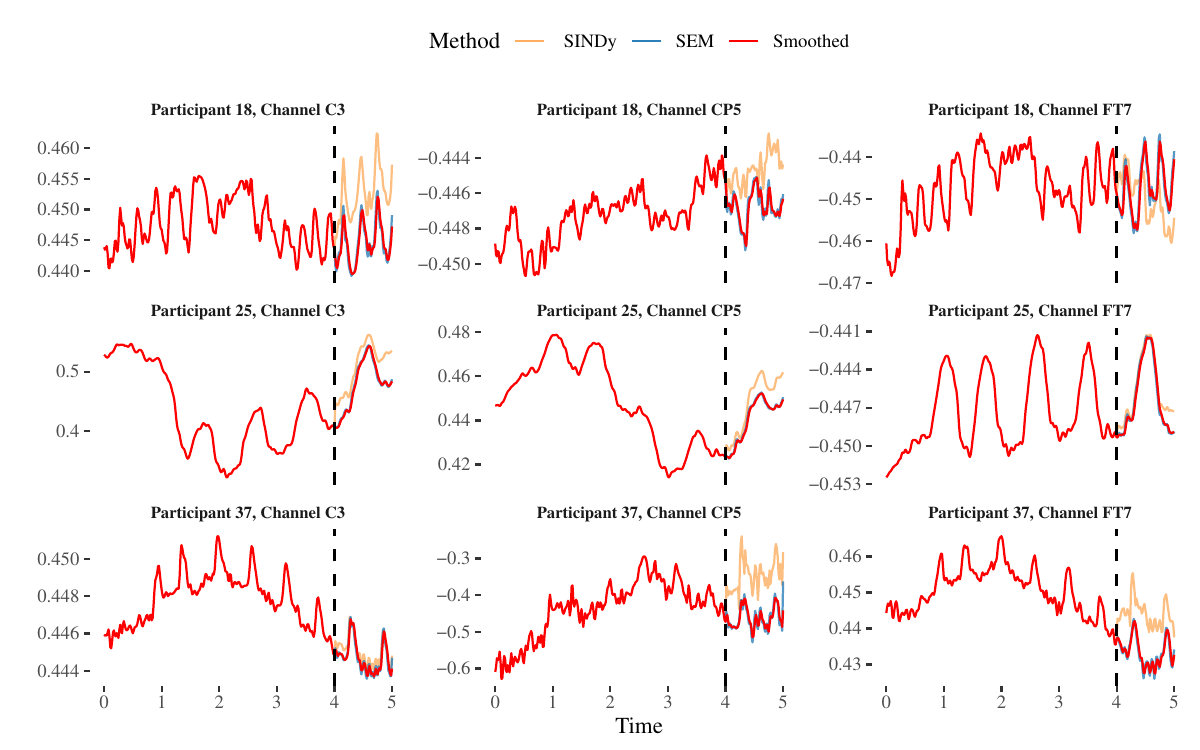}
   \caption{\small 
   Smoothed and predicted dynamics of the eye-blinking task for EEG channels C3, CP5, and FT7 
   from participants \( h = 18, 25, 37 \). 
   The dashed vertical lines mark the transition from the training interval (\( t < 4 \)) 
   to the validation interval (\( t \geq 4 \)).
   }
   \label{fig:data_curve}
\end{figure}

Since the trajectories are densely observed, we apply the RKHS regression~\eqref{eq:estimator_optimization} to the noisy data 
\(\{Y_{ij}^{(h)}; i = 1, \ldots, p,\ j = 1, \ldots, n\}\) 
to recover the smoothed latent trajectories 
\(\bm{X}^{(h)} = ({X}_1^{(h)}, \dots, {X}_p^{(h)})^{\top}\) 
for the \( h \)th participant. 
The denoised trajectory is denoted as \( \hat{\bm{X}}^{(h)} \).
In Figure~\ref{fig:data_curve}, we illustrate the denoised trajectories of several EEG channels for the eye-blinking task, 
where sustained oscillatory patterns are observed in the EEG signals. 
Such oscillations may result from the interplay between inertia and damping, 
and their accurate characterization requires modeling acceleration—hence, a second-order formulation.

Accordingly, we model the data using a second-order nonlinear differential equation:
\begin{align}
\label{dynamic-model}
    \frac{\mathrm{d}^2 X_{i}^{(h)}(t)}{\mathrm{d}t^2} 
    = -\omega_{i1}\frac{\mathrm{d}X_{i}^{(h)}(t)}{\mathrm{d}t} 
    + \bm H^\top(\bm{X}^{(h)}(t)) \bm\beta_i, 
    \quad i = 1, \dots, p,
\end{align}
where \( \bm H(\bm{X}^{(h)}(t)) \) contains candidate basis functions for modeling the dynamics. 
In our analysis, we construct \( \bm H(\bm{X}^{(h)}(t)) \) as in \eqref{h_candidate}, 
using a polynomial basis of order \( P = 2 \). 
Such formulations of ODEs are widely adopted in the literature 
\citep{deco2008dynamic, zhang2017bayesian, brunton2022data}, 
as they naturally accommodate the acceleration and damping effects commonly observed in neural systems.

\paragraph*{Prediction Performance} For each participant and each oculomotor task, we apply SEM with a LASSO penalty to estimate the underlying differential equations from \( \hat{\bm{X}}^{(h)} \). 
We also compare SEM with the benchmark method SINDy, which is widely used in brain dynamics and network discovery~\citep{jansen1995electroencephalogram, friston2003dynamic, zhang2020bayesian}.

To evaluate the performance of the two methods, we partition each smoothed trajectory for the oculomotor task into a training interval 
(\( t \in [0, 4) \)) and a validation interval (\( t \in [4, 5] \)). 
We then compute the predicted values \( \hat{X}_{i}^{(h),\mathrm{pre}}(t_j) \), obtained by solving the estimated differential system with the initial condition 
\( \hat{\bm{X}}^{(h)}(t_j - \Delta t) \), where \( \Delta t = 0.015625 \) seconds.

Figure~\ref{fig:data_curve} illustrates the predicted trajectories \( \hat{X}_{i}^{(h),\mathrm{pre}} \) from SINDy and SEM, 
focusing on the eye-blinking task from three participants (\( h = 18, 25, 37 \)) across three EEG channels (C3, CP5, and FT7). 
SEM predictions closely align with the smoothed dynamics in the validation interval, 
whereas SINDy exhibits noticeable prediction bias.

\begin{figure}[htbp]
    \centering
    \includegraphics[width=0.75\linewidth]{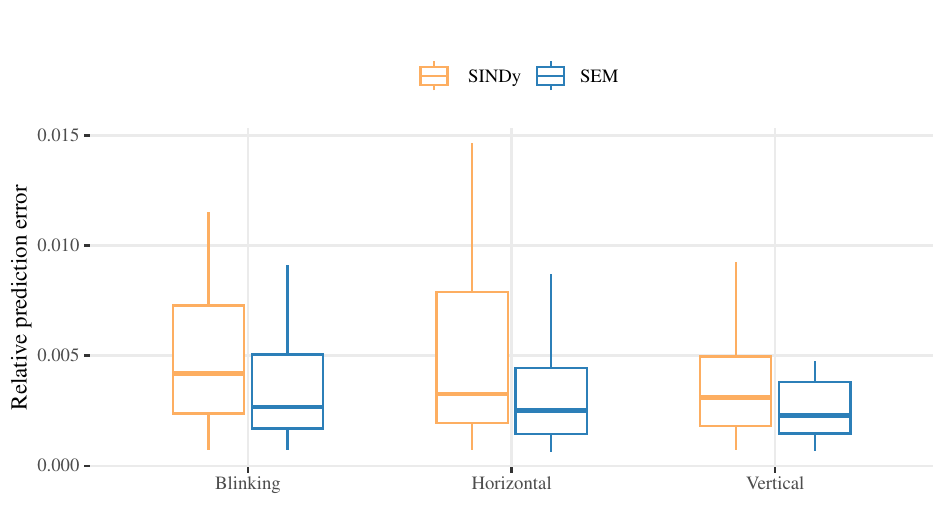}
    \caption{\small
    Boxplots of relative prediction errors (RPE) for 52 participants across three oculomotor tasks: 
    eye blinking (blinking), horizontal eye movement (horizontal), and vertical eye movement (vertical). 
    The RPE is evaluated on the held-out validation interval \([4,5]\).
    }
    \label{fig:data_RER}
\end{figure}

We define the one-step-ahead relative prediction error (RPE) for the \( h \)th participant as:
\[
\mathrm{RPE}_h := 
\frac{1}{p}\sqrt{
\sum_{i=1}^p 
\frac{\sum_{j=1}^T 
\left( \hat{X}_{i}^{(h),\mathrm{pre}}(t_j) - \hat{X}_{i}^{(h)}(t_j) \right)^2}
{\sum_{j=1}^T \hat{X}_{i}^{(h)}(t_j)^2}
},
\]
where \( \{t_j\}_{j=1}^T \) are 512 equally spaced validation points in \( [4, 5] \).
Figure~\ref{fig:data_RER} reports the values of \( \mathrm{RPE}_h \) for all participants under the eye-blinking, horizontal eye movement, and vertical eye movement tasks. 
The results show that SEM consistently outperforms SINDy across all tasks. 
The inferior performance of SINDy may stem from errors associated with the estimation of high-order derivatives, 
whereas SEM circumvents derivative estimation, leading to more accurate predictive performance.

\paragraph*{Brain Network Discovery} Based on the equations discovered by SEM, we construct directed brain networks for each participant and each oculomotor task. 
Let \( E = \{e_{i,j}\}_{i,j=1}^p \) denote the adjacency matrix derived from the dynamical model in~\eqref{dynamic-model}, 
where \( e_{i,j} = 1 \) if \( X_j \) contributes to the second-order derivative of \( X_i \), and \( e_{i,j} = 0 \) otherwise.  
For each participant \( h \in \{1, \dots, 52\} \), we estimate an individual-specific adjacency matrix 
\( \hat{E}^{(h)} = \{\hat{e}_{i,j}^{(h)}\}_{i,j=1}^p \) from the discovered dynamics, 
resulting in a total of 52 networks for each task.  

To infer the population-level network, we treat \( \hat{E}^{(h)} \), \( h \in \{1, \dots, 52\} \), 
as repeated observations of \( E \), and assume that each edge \( e_{i,j} \) follows a Bernoulli distribution: 
\( e_{i,j} \sim \mathrm{Bernoulli}(p_{i,j}) \). 
Under this setup, we test for the presence of each edge \( e_{i,j} \) using the one-sided hypothesis test:
\begin{align}\label{test_1}
    H_0: p_{i,j} \leq 0.5 \quad \text{versus} \quad H_1: p_{i,j} > 0.5,
\end{align}
at significance level \( \alpha = 0.05 \). 
We conduct this hypothesis test for all edges \( e_{i,j} \) based on the collection of individual network estimates 
\( \{\hat{E}^{(h)}\}_{h=1}^{52} \).

Note that the above task constitutes a multiple testing problem. 
To control for false discoveries, we apply the Benjamini–Hochberg (BH) procedure \citep{benjamini1995controlling}, 
ensuring that the false discovery rate (FDR) remains below \( 5\% \). 
The resulting significant edges for the resting state and the three oculomotor tasks are presented in 
Figure~\ref{fig:all-tasks} in Appendix~\ref{app-edge}. 
We observe that the three oculomotor tasks exhibit distinct network connectivity patterns, 
whereas no significant directed edges are detected in the resting state.

\begin{figure*}[t]
    \centering
    \includegraphics[width=\textwidth]{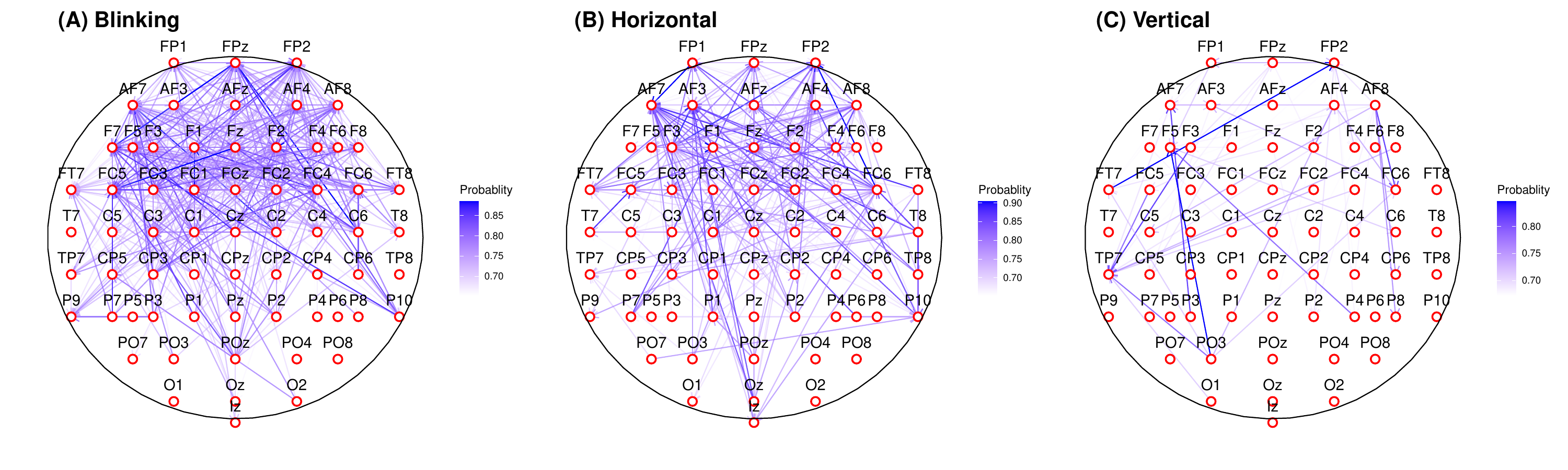}
    \caption{\small 
    Directed brain networks for the eye-blinking, horizontal eye movement, and vertical eye movement tasks. 
    Edge color represents the empirical probability of occurrence of each edge across participants.
    } 
    \label{fig:net_tasks}
\end{figure*}

To further investigate task-specific brain networks, we construct a second type of network that identifies edges 
that are more active during oculomotor tasks than during the resting state. 
Specifically, let \( {E}^{0} = \{{e}^{0}_{i,j}\}_{i,j=1}^p \) and \( {E}^1 = \{{e}_{i,j}^1\}_{i,j=1}^p \) 
denote the adjacency matrices under the resting-state condition and a given oculomotor task, respectively. 
For each edge \( (i,j) \), we perform a one-sided Fisher’s exact test to assess the null hypothesis:
\begin{eqnarray*}
     & H_0: \mathbb{P}({e}_{i,j}^1 = 1) \leq \mathbb{P}({e}_{i,j}^{0} = 1),  \\
     & H_1: \mathbb{P}({e}_{i,j}^1 = 1) > \mathbb{P}({e}_{i,j}^{0} = 1),
\end{eqnarray*}
at a significance level of \( \alpha = 0.05 \). 
We apply the Benjamini–Hochberg (BH) procedure to control the false discovery rate at \( 5\% \). 
Edges found to be significant in both the original test in \eqref{test_1} 
and the Fisher’s exact test above are retained to construct a refined task-specific network. 
The resulting networks for the three oculomotor tasks are shown in Figure~\ref{fig:net_tasks}.

We compute several node-level metrics—out-degree, in-degree, betweenness centrality, and closeness centrality \citep{rubinov2010complex}—to identify active brain regions in Figure~\ref{fig:net_tasks}. 
Specifically, out-degree and in-degree measure a node’s tendency to initiate or receive information, respectively. 
Betweenness centrality quantifies how often a node serves as a bridge along communication paths, 
while closeness centrality evaluates how efficiently a node can reach all other nodes in the network. 
The computed metrics for the three oculomotor tasks are provided in Appendix~\ref{app-edge}.

In the eye-blinking task, significant brain connectivity is concentrated in the prefrontal region 
(see Figure~\ref{fig:net_tasks}~(A)). Within this region, channels FP2, F7, F5, and FPz exhibit high betweenness centrality 
(see Figure~\ref{fig:graph-count-blink}~(A) in Appendix~\ref{app-edge}). 
Additionally, FP2 and F5 show high in-degree values 
(see Figure~\ref{fig:graph-count-blink}~(C) in Appendix~\ref{app-edge}), 
with FP2 also exhibiting a relatively high out-degree 
(see Figure~\ref{fig:graph-count-blink}~(D) in Appendix~\ref{app-edge}). 
For the horizontal eye movement task, we similarly observe strong connectivity in the prefrontal region 
(see Figure~\ref{fig:net_tasks}~(B)). 
The prefrontal channels F3, AF4, AF3, and AF8 display high betweenness centrality 
(see Figure~\ref{fig:graph-count-left}~(A) in Appendix~\ref{app-edge}), 
while AF7, FP2, and AF3 exhibit high in-degree values 
(see Figure~\ref{fig:graph-count-left}~(C) in Appendix~\ref{app-edge}).

In the vertical eye movement task, we observe a posterior-to-anterior connectivity pattern 
(see Figure~\ref{fig:net_tasks}~(C)), characterized by a directed link from PO3 to F5. 
Channels FP2, FT7, and AF4 exhibit high betweenness centrality 
(see Figure~\ref{fig:graph-count-up}~(A) in Appendix~\ref{app-edge}), 
with FT7 also showing a high in-degree 
(see Figure~\ref{fig:graph-count-up}~(C) in Appendix~\ref{app-edge}).

Based on the above results, we identify prominent neural activity in the frontal cortical region, 
particularly involving channels such as FP2, F7, F5, FPz, AF3, and AF8—areas known to be associated 
with oculomotor control \citep{kato2003functional, van2012neural}. 
Among them, channel FP2 consistently exhibits high centrality across all three tasks, 
indicating its potential role as a central hub for oculomotor coordination. 
This channel is located over the right prefrontal cortex, 
which has been linked to executive functions and voluntary saccadic regulation 
\citep{funahashi2014saccade}.

For horizontal eye movements, we find that high-centrality nodes display a bilaterally symmetric 
distribution (e.g., AF4, AF8 $\leftrightarrow$ AF3, AF7), suggesting that the task may involve 
interhemispheric coordination mechanisms. 
In contrast, vertical eye movements exhibit a distinct anterior–posterior connectivity pattern 
(e.g., PO3 $\to$ F5), with weaker bilateral connectivity compared to the horizontal task. 
As noted in the literature, vertical eye movements may rely more heavily on intrahemispheric 
signal transmission—such as anterior–posterior pathways—during oculomotor control 
\citep{horn2011anatomy, leigh2015neurology}. 
Our findings support this phenomenon, revealing significant connectivity differences 
between horizontal and vertical eye movements.

\section{Discussion}
\label{Sec: discuss}
In this article, we propose a novel sparse equation matching (SEM) framework for equation discovery 
in general-order dynamical systems. 
Our approach unifies a broad class of existing methods through Green’s function representations, 
enabling derivative-free estimation of both general-order differential operators 
and driving functions from dynamic data. 
The effectiveness of SEM is demonstrated through extensive simulations, 
highlighting its superior performance over derivative-based methods 
in discovering general-order dynamics.
We further apply SEM to EEG data collected during multiple oculomotor tasks, 
uncovering meaningful and interpretable brain connectivity patterns 
based on the second-order signals of the data. 

The current SEM framework has some limitations. 
First, our method focuses on sparse equation matching using a predefined set of candidate basis functions. 
Alternative approaches—such as Gaussian processes \citep{heinonen2018learning}, 
RKHS-based models \citep{bobade2019adaptive, dai2022kernel}, 
symbolic regression \citep{qian2022d}, 
and neural networks \citep{chen2018neural}—
could be integrated into our framework to explore broader structural forms of governing equations.
Second, SEM relies on a pre-estimation step for trajectory recovery, 
making it a two-step procedure for differential equation learning. 
This approach is well-suited for densely observed data such as the EEG dataset in our analysis, 
where trajectory pre-estimates can be accurately obtained. 
However, in settings with sparse or irregular observations, 
the pre-estimation step may introduce additional errors 
that propagate into the subsequent equation discovery. 
Methods that jointly estimate trajectories and underlying differential equations, 
such as \citet{ramsay2007parameter, raissi2019physics, yang2021inference}, 
may offer more robust alternatives in such cases. 
Extending SEM to incorporate joint estimation is an important direction for future research.

\bibliographystyle{apalike}
\bibliography{ref}

@article{shao2025ordinary,
  title={Ordinary differential equation models for a collection of discretized functions},
  author={Shao, Lingxuan and Yao, Fang},
  journal={Journal of the Royal Statistical Society Series B: Statistical Methodology},
  pages={qkaf036},
  year={2025}, 
  volume={ },
  number={ },
  publisher={Oxford University Press UK}
}

@article{tan2024functional,
      title={Functional Singular Value Decomposition}, 
      author={Jianbin Tan and Pixu Shi and Anru R. Zhang},
      year={2025},
      journal={arXiv preprint arXiv:2410.03619},
}

@book{hsing2015theoretical,
  title={Theoretical foundations of functional data analysis, with an introduction to linear operators},
  author={Hsing, Tailen and Eubank, Randall},
  volume={997},
  year={2015},
  publisher={John Wiley \& Sons},
  address = {Chichester, UK}
}

@article{tan2022transmission,
  title={Transmission roles of symptomatic and asymptomatic COVID-19 cases: a modelling study},
  author={Tan, Jianbin and Ge, Yang and Martinez, Leonardo and Sun, Jimin and Li, Changwei and Westbrook, Adrianna and Chen, Enfu and Pan, Jinren and Li, Yang and Cheng, Wei and others},
  journal={Epidemiology \& Infection},
  volume={150},
  pages={e171},
  number={ },
  year={2022},
  publisher={Cambridge University Press}
}

@article{tan2024green,
  title={Green’s matching: an efficient approach to parameter estimation in complex dynamic systems},
  author={Tan, Jianbin and Zhang, Guoyu and Wang, Xueqin and Huang, Hui and Yao, Fang},
  journal={Journal of the Royal Statistical Society Series B: Statistical Methodology},
  pages={qkae031},
  year={2024},
  volume={ },
  number={ },
  publisher={Oxford University Press UK}
}

@article{dattner2015optimal,
  title={Optimal rate of direct estimators in systems of ordinary differential equations linear in functions of the parameters},
  author={Dattner, Itai and Klaassen, Chris AJ},
  journal = {Electronic Journal of Statistics},
  year={2015},
  volume = {9},
  number = {2},
  publisher = {Institute of Mathematical Statistics and Bernoulli Society},
  pages = {1939 -- 1973},
}

@article{wu2014sparse,
  title={Sparse additive ordinary differential equations for dynamic gene regulatory network modeling},
  author={Wu, Hulin and Lu, Tao and Xue, Hongqi and Liang, Hua},
  journal={Journal of the American Statistical Association},
  volume={109},
  number={506},
  pages={700--716},
  year={2014},
  publisher={Taylor \& Francis}
}

@article{chen2017network,
  title={Network reconstruction from high-dimensional ordinary differential equations},
  author={Chen, Shizhe and Shojaie, Ali and Witten, Daniela M},
  journal={Journal of the American Statistical Association},
  volume={112},
  number={520},
  pages={1697--1707},
  year={2017},
  publisher={Taylor \& Francis}
}

@article{dai2022kernel,
  title={Kernel ordinary differential equations},
  author={Dai, Xiaowu and Li, Lexin},
  journal={Journal of the American Statistical Association},
  volume={117},
  number={540},
  pages={1711--1725},
  year={2022},
  publisher={Taylor \& Francis}
}

@inproceedings{chen2018neural,
 author = {Chen, Ricky T. Q. and Rubanova, Yulia and Bettencourt, Jesse and Duvenaud, David K},
 booktitle = {Advances in Neural Information Processing Systems},
 pages = {},
 publisher = {Curran Associates, Inc.},
 title = {Neural Ordinary Differential Equations},
 volume = {31},
 year = {2018}
}

@book{ramsay2017dynamic,
  title={Dynamic Data Analysis},
  author={Ramsay, James and Hooker, Giles},
  year={2017},
  publisher={Springer},
  address={New York, NY},
}

@article{zhang2017bayesian,
  title={Bayesian inference of high-dimensional, cluster-structured ordinary differential equation models with applications to brain connectivity studies},
  author={Zhang, Tingting and Yin, Qiannan and Caffo, Brian and Sun, Yinge and Boatman-Reich, Dana},
  volume = {11},
  number = {2},
  publisher = {Institute of Mathematical Statistics},
  pages = {868 -- 897},
  journal = {The Annals of Applied Statistcs},
  year={2017}
}

@article{zhang2020bayesian,
  title={Bayesian inference of a directional brain network model for intracranial EEG data},
  author={Zhang, Tingting and Sun, Yinge and Li, Huazhang and Yan, Guofen and Tanabe, Seiji and Miao, Ruizhong and Wang, Yaotian and Caffo, Brian S and Quigg, Mark S},
  journal={Computational Statistics \& Data Analysis},
  volume={144},
  pages={106847},
  year={2020},
  publisher={Elsevier}
}

@article{tian2021effects,
  title={The effects of stringent and mild interventions for coronavirus pandemic},
  author={Tian, Ting and Tan, Jianbin and Luo, Wenxiang and Jiang, Yukang and Chen, Minqiong and Yang, Songpan and Wen, Canhong and Pan, Wenliang and Wang, Xueqin},
  journal={Journal of the American Statistical Association},
  volume={116},
  number={534},
  pages={481--491},
  year={2021},
  publisher={Taylor \& Francis}
}

@article{tan2023age,
  title={Age-related model for estimating the symptomatic and asymptomatic transmissibility of COVID-19 patients},
  author={Tan, Jianbin and Shen, Ye and Ge, Yang and Martinez, Leonardo and Huang, Hui},
  journal={Biometrics},
  volume={79},
  number={3},
  pages={2525--2536},
  year={2023},
  publisher={Oxford University Press}
}

@article{brunton2016discovering,
  title={Discovering governing equations from data by sparse identification of nonlinear dynamical systems},
  author={Brunton, Steven L and Proctor, Joshua L and Kutz, J Nathan},
  journal={Proceedings of the National Academy of Sciences},
  volume={113},
  number={15},
  pages={3932--3937},
  year={2016},
  publisher={National Acad Sciences}
}

@article{champion2019data,
  title={Data-driven discovery of coordinates and governing equations},
  author={Champion, Kathleen and Lusch, Bethany and Kutz, J Nathan and Brunton, Steven L},
  journal={Proceedings of the National Academy of Sciences},
  volume={116},
  number={45},
  pages={22445--22451},
  year={2019},
  publisher={National Academy of Sciences}
}

@article{kaiser2018sparse,
  title={Sparse identification of nonlinear dynamics for model predictive control in the low-data limit},
  author={Kaiser, Eurika and Kutz, J Nathan and Brunton, Steven L},
  journal={Proceedings of the Royal Society A},
  volume={474},
  number={2219},
  pages={20180335},
  year={2018},
  publisher={The Royal Society Publishing}
}

@article{fasel2022ensemble,
  title={Ensemble-SINDy: Robust sparse model discovery in the low-data, high-noise limit, with active learning and control},
  author={Fasel, Urban and Kutz, J Nathan and Brunton, Bingni W and Brunton, Steven L},
  journal={Proceedings of the Royal Society A: Mathematical, Physical and Engineering Sciences},
  volume={478},
  number={2260},
  pages={20210904},
  year={2022},
  publisher={The Royal Society}
}

@article{egan2024automatically,
  title={Automatically discovering ordinary differential equations from data with sparse regression},
  author={Egan, Kevin and Li, Weizhen and Carvalho, Rui},
  journal={Communications Physics},
  volume={7},
  number={1},
  pages={20},
  year={2024},
  publisher={Nature Publishing Group UK London}
}

@article{messenger2021weak,
  title={Weak SINDy for partial differential equations},
  author={Messenger, Daniel A and Bortz, David M},
  journal={Journal of Computational Physics},
  volume={443},
  pages={110525},
  year={2021},
  publisher={Elsevier}
}

@inproceedings{qian2022d,
  title={D-code: Discovering closed-form odes from observed trajectories},
  author={Qian, Zhaozhi and Kacprzyk, Krzysztof and van der Schaar, Mihaela},
  booktitle={International Conference on Learning Representations},
  year={2022}
}

@article{zou2005regularization,
  title={Regularization and variable selection via the elastic net},
  author={Zou, Hui and Hastie, Trevor},
  journal={Journal of the Royal Statistical Society Series B: Statistical Methodology},
  volume={67},
  number={2},
  pages={301--320},
  year={2005},
  publisher={Oxford University Press}
}

@article{zhu2020polynomial,
  title={A polynomial algorithm for best-subset selection problem},
  author={Zhu, Junxian and Wen, Canhong and Zhu, Jin and Zhang, Heping and Wang, Xueqin},
  journal={Proceedings of the National Academy of Sciences},
  volume={117},
  number={52},
  pages={33117--33123},
  year={2020},
  publisher={National Acad Sciences}
}

@article{yuan2006model,
  title={Model selection and estimation in regression with grouped variables},
  author={Yuan, Ming and Lin, Yi},
  journal={Journal of the Royal Statistical Society Series B: Statistical Methodology},
  volume={68},
  number={1},
  pages={49--67},
  year={2006},
  publisher={Oxford University Press}
}

@article{schaeffer2017sparse,
  title={Sparse model selection via integral terms},
  author={Schaeffer, Hayden and McCalla, Scott G},
  journal={Physical Review E},
  volume={96},
  number={2},
  pages={023302},
  year={2017},
  publisher={APS}
}

@book{gu2013smoothing,
  title={Smoothing spline ANOVA models},
  author={Gu, Chong and Gu, Chong},
  volume={297},
  year={2013},
  publisher={Springer},
  address = {New York}
}

@book{hastie2017elements,
  title={The elements of statistical learning: data mining, inference, and prediction},
  author={Hastie, Trevor and Tibshirani, Robert and Friedman, Jerome},
  year={2017},
  publisher={Springer},
  address = {New York}
}

@book{thompson2002nonlinear,
  title={Nonlinear dynamics and chaos},
  author={Thompson, John Michael Tutill and Stewart, H Bruce},
  year={2002},
  publisher={John Wiley \& Sons},
  address  = {New York}
}

@book{hairergeometric,
  title={Geometric Numerical Integration: Structure-Preserving Algorithms for Ordinary Differential Equations},
  year = {2006},
  author={Hairer, E and Lubich, C and Wanner, G},
  publisher={Springer-Verlag},
  address = {Berlin}
}

@book{sporns2016networks,
  title={Networks of the Brain},
  author={Sporns, Olaf},
  year={2016},
  publisher={MIT press},
  address = {Cambridge, MA}
}

@article{cho2017eeg,
  title={EEG datasets for motor imagery brain--computer interface},
  author={Cho, Hohyun and Ahn, Minkyu and Ahn, Sangtae and Kwon, Moonyoung and Jun, Sung Chan},
  journal={GigaScience},
  volume={6},
  number={7},
  pages={gix034},
  year={2017},
  publisher={Oxford University Press}
}

@book{duffy2015green,
  title={Green's functions with applications},
  author={Duffy, Dean G},
  year={2015},
  publisher={Chapman and Hall/CRC},
  address = {New York}
}

@article{kato2003functional,
  title={Functional MRI of brain activation evoked by intentional eye blinking},
  author={Kato, Makoto and Miyauchi, Satoru},
  journal={Neuroimage},
  volume={18},
  number={3},
  pages={749--759},
  year={2003},
  publisher={Elsevier}
}

@article{van2012neural,
  title={Neural control of voluntary eye closure: a case study and an fMRI investigation of blinking and winking},
  author={van Koningsbruggen, Martijn G and Peelen, Marius V and Davies, Eilir and Rafal, Robert D},
  journal={Behavioural Neurology},
  volume={25},
  number={2},
  pages={103--109},
  year={2012},
  publisher={Wiley Online Library}
}

@article{lu2021learning,
  title={Learning nonlinear operators via DeepONet based on the universal approximation theorem of operators},
  author={Lu, Lu and Jin, Pengzhan and Pang, Guofei and Zhang, Zhongqiang and Karniadakis, George Em},
  journal={Nature Machine Intelligence},
  volume={3},
  number={3},
  pages={218--229},
  year={2021},
  publisher={Nature Publishing Group UK London}
}

@inproceedings{heinonen2018learning,
  title={Learning unknown ODE models with Gaussian processes},
  author={Heinonen, Markus and Yildiz, Cagatay and Mannerstr{\"o}m, Henrik and Intosalmi, Jukka and L{\"a}hdesm{\"a}ki, Harri},
  booktitle={International Conference on Machine Learning},
  pages={1959--1968},
  year={2018},
  organization={PMLR}
}

@book{brunton2022data,
  title={Data-driven science and engineering: Machine learning, dynamical systems, and control},
  author={Brunton, Steven L and Kutz, J Nathan},
  year={2022},
  publisher={Cambridge University Press},
  address = {Cambridge, UK}
}

@book{wahba1990spline,
  title={Spline models for observational data},
  author={Wahba, Grace},
  year={1990},
  publisher={SIAM},
  address={Philadelphia}
}

@article{rubinov2010complex,
  title={Complex network measures of brain connectivity: uses and interpretations},
  author={Rubinov, Mikail and Sporns, Olaf},
  journal={Neuroimage},
  volume={52},
  number={3},
  pages={1059--1069},
  year={2010},
  publisher={Elsevier}
}

@article{benjamini1995controlling,
  title={Controlling the false discovery rate: a practical and powerful approach to multiple testing},
  author={Benjamini, Yoav and Hochberg, Yosef},
  journal={Journal of the Royal Statistical Society: Series B (Methodological)},
  volume={57},
  number={1},
  pages={289--300},
  year={1995},
  publisher={Wiley Online Library}
}

@article{friston2003dynamic,
  title={Dynamic causal modelling},
  author={Friston, Karl J and Harrison, Lee and Penny, Will},
  journal={Neuroimage},
  volume={19},
  number={4},
  pages={1273--1302},
  year={2003},
  publisher={Elsevier}
}

@article{Friston2009,
author = {Kiebel, Stefan J. and Garrido, Marta I. and Moran, Rosalyn and Chen, Chun-Chuan and Friston, Karl J.},
title = {Dynamic causal modeling for EEG and MEG},
journal = {Human Brain Mapping},
volume = {30},
number = {6},
pages = {1866-1876},
year = {2009}
}

@article{Friston2011review,
author = {Friston, Karl J},
title = {Functional and effective connectivity: a review},
journal = {Brain Connectivity},
volume = {1},
number={1},
pages = {13-36},
year = {2011},
publisher = {Mary Ann Liebert, Inc.}
}

@article{deco2008dynamic,
  title={The dynamic brain: from spiking neurons to neural masses and cortical fields},
  author={Deco, Gustavo and Jirsa, Viktor K and Robinson, Peter A and Breakspear, Michael and Friston, Karl},
  journal={PLoS Computational Biology},
  volume={4},
  number={8},
  pages={e1000092},
  year={2008},
  publisher={Public Library of Science San Francisco, USA}
}

@article{jansen1995electroencephalogram,
  title={Electroencephalogram and visual evoked potential generation in a mathematical model of coupled cortical columns},
  author={Jansen, Ben H and Rit, Vincent G},
  journal={Biological Cybernetics},
  volume={73},
  number={4},
  pages={357--366},
  year={1995},
  publisher={Springer}
}

@article{yang2021inference,
  title={Inference of dynamic systems from noisy and sparse data via manifold-constrained Gaussian processes},
  author={Yang, Shihao and Wong, Samuel WK and Kou, SC},
  journal={Proceedings of the National Academy of Sciences},
  volume={118},
  number={15},
  pages={e2020397118},
  year={2021},
  publisher={National Academy of Sciences}
}

@article{raissi2019physics,
  title={Physics-informed neural networks: A deep learning framework for solving forward and inverse problems involving nonlinear partial differential equations},
  author={Raissi, Maziar and Perdikaris, Paris and Karniadakis, George E},
  journal={Journal of Computational Physics},
  volume={378},
  pages={686--707},
  year={2019},
  publisher={Elsevier}
}

@article{bobade2019adaptive,
  title={Adaptive estimation for nonlinear systems using reproducing kernel Hilbert spaces},
  author={Bobade, Parag and Majumdar, Suprotim and Pereira, Savio and Kurdila, Andrew J and Ferris, John B},
  journal={Advances in Computational Mathematics},
  volume={45},
  pages={869--896},
  year={2019},
  publisher={Springer}
}

@article{funahashi2014saccade,
  title={Saccade-related activity in the prefrontal cortex: its role in eye movement control and cognitive functions},
  author={Funahashi, Shintaro},
  journal={Frontiers in Integrative Neuroscience},
  volume={8},
  pages={54},
  year={2014},
  publisher={Frontiers Media SA}
}

@article{lu2019nonparametric,
  title={Nonparametric inference of interaction laws in systems of agents from trajectory data},
  author={Lu, Fei and Zhong, Ming and Tang, Sui and Maggioni, Mauro},
  journal={Proceedings of the National Academy of Sciences},
  volume={116},
  number={29},
  pages={14424--14433},
  year={2019},
  publisher={National Academy of Sciences}
}

@book{niedermeyer2005electroencephalography,
  title={Electroencephalography: basic principles, clinical applications, and related fields},
  author={Niedermeyer, Ernst and da Silva, FH Lopes},
  year={2005},
  publisher={Lippincott Williams \& Wilkins},
  address={Philadelphia, PA}
}

@book{shumway2000time,
  title={Time series analysis and its applications},
  author={Shumway, Robert H and Stoffer, David S and Stoffer, David S},
  volume={3},
  year={2000},
  publisher={Springer},
  address={New York}
}

@article{ramsay2007parameter,
  title={Parameter estimation for differential equations: a generalized smoothing approach},
  author={Ramsay, Jim O and Hooker, Giles and Campbell, David and Cao, Jiguo},
  journal={Journal of the Royal Statistical Society Series B: Statistical Methodology},
  volume={69},
  number={5},
  pages={741--796},
  year={2007},
  publisher={Oxford University Press}
}

@incollection{lorenz2017deterministic,
  title={Deterministic Nonperiodic Flow 1},
  author={Lorenz, Edward N},
  booktitle={Universality in Chaos, 2nd edition},
  pages={367--378},
  year={2017},
  publisher={Routledge},
  address={New York}
}

@book{meirovitch2010fundamentals,
  title={Fundamentals of Vibrations},
  author={Meirovitch, Leonard},
  year={2010},
  publisher={Waveland Press},
  address={Long Grove, IL}
}

@book{leigh2015neurology,
  title={The neurology of eye movements},
  author={Leigh, R John and Zee, David S},
  year={2015},
  publisher={Oxford university press},
  address={Oxford, UK}
}

@article{horn2011anatomy,
  title={The anatomy and physiology of the ocular motor system},
  author={Horn, Anja KE and Leigh, R John},
  journal={Handbook of Clinical Neurology},
  volume={102},
  pages={21--69},
  year={2011},
  publisher={Elsevier}
}

@article{owens2023data,
  title={Data-driven discovery of governing equations for coarse-grained heterogeneous network dynamics},
  author={Owens, Katherine and Kutz, J Nathan},
  journal={SIAM Journal on Applied Dynamical Systems},
  volume={22},
  number={3},
  pages={2601--2623},
  year={2023},
  publisher={SIAM}
}

@article{stein2014enso,
  title={ENSO seasonal synchronization theory},
  author={Stein, Karl and Timmermann, Axel and Schneider, Niklas and Jin, Fei-Fei and Stuecker, Malte F},
  journal={Journal of Climate},
  volume={27},
  number={14},
  pages={5285--5310},
  year={2014},
  publisher = {American Meteorological Society}
}

@article{carrillo2017review,
  title={A review on attractive--repulsive hydrodynamics for consensus in collective behavior},
  author={Carrillo, Jos{\'e} A and Choi, Young-Pil and Perez, Sergio P},
  journal={Active Particles, Volume 1: Advances in Theory, Models, and Applications},
  pages={259--298},
  year={2017},
  volume={1},
  publisher={Springer}
}

@Manual{R-fdapace,
  title  = {fdapace: Functional Data Analysis and Empirical Dynamics},
  author = {Lingzhou Zhou and Shuo Lin and Jiahui Yang and Hans-Georg Müller},
  year   = {2022},
  note   = {R package version 0.5.9},
  url    = {https://CRAN.R-project.org/package=fdapace},
}

@article{xiao2019asymptotic,
  title        = {Asymptotic theory of penalized splines},
  author       = {Xiao, Luo},
  journal      = {Electronic Journal of Statistics},
  volume       = {13},
  number       = {1},
  pages        = {747--794},
  year         = {2019},
  publisher    = {Institute of Mathematical Statistics and Bernoulli Society},
}

@article{shang2013local,
  title={Local and global asymptotic inference in smoothing spline models},
  author={Shang, Zuofeng and Cheng, Guang},
  journal={The Annals of Statistics},
  volume={41},
  number={5},
  pages={2608--2638},
  year={2013},
  publisher={Institute of Mathematical Statistics},
}

@article{stone1980optimal,
  title={Optimal rates of convergence for nonparametric estimators},
  author={Stone, Charles J.},
  journal={Annals of Statistics},
  volume={8},
  number={6},
  pages={1348--1360},
  year={1980},
  publisher = {Institute of Mathematical Statistics}
}

@article{zhou2000derivative,
  title={On derivative estimation in spline regression},
  author={Zhou, Shanggang and Wolfe, Douglas A.},
  journal={Statistica Sinica},
  volume={10},
  number={1},
  pages={93--108},
  year={2000},
  publisher = {Institute of Mathematical Statistics}

}

@article{antwi2022naive,
  title={Naive penalized spline estimators of derivatives achieve optimal rates of convergence},
  author={Antwi, Bright and Staudenmayer, John},
  journal={arXiv preprint arXiv:2206.06849},
  year={2022}
}

\appendix

\clearpage
\section{Technical Proof}

\subsection{Proof of Theorem \ref{theo:inte}}
\begin{proof}
We divide the proof into two parts based on the value of \( k \).

When \( k = 0 \), by definition, \( G^0(t, s) = \delta(t - s) \), and the kernel term \( g_0(t) \equiv 0 \). Then the integral equation \eqref{match-k} reduces to:
\[
\frac{\mathrm{d}^K X_i}{\mathrm{d}t^K}(t) + \sum_{l=1}^{K-1} \omega_{il} \frac{\mathrm{d}^l X_i}{\mathrm{d}t^l}(t) = f_i(\bm X(t),t),
\]
which is exactly the original differential equation \eqref{ode}.

Then, let \( k \geq 1 \) be fixed. Applying the Green's function \( G^k \) to both sides of \eqref{ode}, we obtain:
\begin{align}
\label{eq:step1}
&\int_0^C G^k(t, s) \left( \frac{\mathrm{d}^K X_i}{\mathrm{d}t^K}(s) + \sum_{l=1}^{K-1} \omega_{il} \frac{\mathrm{d}^l X_i}{\mathrm{d}t^l}(s) \right) \mathrm{d}s= \int_0^C G^k(t, s) f_i(\bm X(s),s)\, \mathrm{d}s.
\end{align}

By the definition of $G^k(t,s)$ in \eqref{standard-green},  we repeatedly apply integration by parts and obtain, for \( l \geq k \),
\begin{align*}
    &\int_0^C G^k(t,s) \frac{\mathrm{d}^{l}X_i}{\mathrm{d}s^{l}}(s)\, \mathrm{d}s\\
    &=\int_0^C\frac{(t - s)^{k - 1}}{(k - 1)!} \, \mathbb{I}(s\leq t) \frac{\mathrm{d}^{l}X_i}{\mathrm{d}t^{l}}(s)\mathrm d s\\
    &=- \frac{t^{k-1}}{(k-1)!}\frac{\mathrm{d}^{l-1}X_i}{\mathrm{d}t^{l-1}}(0) +\int_0^C G^{k-1}(t,s)\frac{\mathrm{d}^{l-1}X_i}{\mathrm{d}t^{l-1}}(s)\mathrm ds\\
    &=\cdots=- \sum_{m=1}^{k}\frac{\mathrm{d}^{l-m}X_i}{\mathrm{d}t^{l-m}}(0)\frac{t^{k-m}}{(k-m)!}+\frac{\mathrm{d}^{l-k}X_i}{\mathrm{d}t^{l-k}}(t),\\
     &= \frac{\mathrm{d}^{l-k} X_i}{\mathrm{d}t^{l-k}}(t) + g_{ik}^{(l)}(t),
\end{align*}
where the the correction term  $g_{ik}^{(l)}(t) \in \mathrm{Ker}\left(\frac{\mathrm d^k}{\mathrm{d}t^k}\right)$.

For \( l < k \), we have:
\[
\int_0^C G^k(t,s) \frac{\mathrm{d}^{l}X_i}{\mathrm{d}s^{l}}(s)\, \mathrm{d}s 
= \int_0^C G^{k-l}(t,s) X_i(s)\, \mathrm{d}s + g_{ik}^{(l)}(t),
\]
where \(g_{ik}^{(l)}(t) \in \mathrm{Ker}\left( \frac{\mathrm{d}^k}{\mathrm{d}t^k}\right)\).
Applying this to each term in \eqref{eq:step1}, we obtain:

\begin{itemize}
    \item For the main term:
    \[
    \int_0^C G^k(t,s) \frac{\mathrm{d}^K X_i}{\mathrm{d}s^K}(s)\, \mathrm{d}s = \frac{\mathrm{d}^{K-k} X_i}{\mathrm{d}t^{K-k}}(t) + g_{ik}^{(K)}(t).
    \]

    \item For \( l \in \{k+1, \ldots, K-1\} \):
    \[
    \int_0^C G^k(t,s) \frac{\mathrm{d}^l X_i}{\mathrm{d}s^l}(s)\, \mathrm{d}s = \frac{\mathrm{d}^{l-k} X_i}{\mathrm{d}t^{l-k}}(t) + g_{ik}^{(l)}(t).
    \]

    \item For \( l \in \{1, \ldots, \min(k, K-1)\} \):
    \[
    \int_0^C G^k(t,s) \frac{\mathrm{d}^l X_i}{\mathrm{d}s^l}(s)\, \mathrm{d}s = \int_0^C G^{k-l}(t,s) X_i(s)\, \mathrm{d}s + g_{ik}^{(l)}(t).
    \]
\end{itemize}

Substituting all terms into equation~\eqref{eq:step1}, we  we arrive at:
\begin{align*}
  & \frac{\mathrm{d}^{K-k}X_i}{\mathrm{d}t^{K-k}}(t) 
    + \sum_{l = k+1}^{K-1} \omega_{il} \frac{\mathrm{d}^{l-k} X_i}{\mathrm{d}t^{l-k}}(t) 
    + \sum_{l = 1}^{\min(k, K-1)} \omega_{il} \cdot\\
    &\;\;\int_0^C G^{k - l}(t, s) X_i(s)\, \mathrm{d}s = g_{ik}(t) + \int_0^C G^k(t,s) f_i(\bm X(s),s)\, \mathrm{d}s,
\end{align*}
where \( g_{ik}(t) := g_{ik}^{(K)}(t) + \sum_{l=1}^{K-1} \omega_{il} g_{ik}^{(l)}(t) \in  \mathrm{Ker}\left(\frac{\mathrm{d}^k}{\mathrm{d}t^k}\right) \).

Specially, when \( k = 0 \), \( \min(k, K-1) = 0 \), so the third summation vanishes.
When \( k \geq K-1 \), the second summation over \( l = k+1, \ldots, K-1 \) becomes empty and thus is omitted.
When \( k \geq K \), the first term \( \frac{\mathrm{d}^{K-k} X_i}{\mathrm{d}t^{K-k}}(t) \) is ill-defined, so we it by its integral counterpart:
  \(    \int_0^C G^{k-K}(t,s) X_i(s)\, \mathrm{d}s.\)

This completes the proof.
\end{proof}

\subsection{Proof of Theorem \ref{thm-optimizer}}
\label{app-ide}
\begin{proof}
    For \[\widetilde G^k(t,s)=G^k(t,s) + \bm\Phi_k^\top(t)\,\bm v_k(s),\;k=1,2,\ldots,K,\] we have the following identities:
\begin{align*}
    \int_0^C\tilde{G}^k(t,s)X_i(s)\mathrm{d}s=\int_0^CG^{k}(t,s)X_i(s)\mathrm{d}s+
\bm\Phi_k^\top(t)\int_0^C\bm{v}_k(s)X_i(s)\mathrm{d}s,\; 1\leq k \leq K-1 \\
\int_0^C\tilde{G}^K(t,s)\bm H^\top(\bm{X}(s))\mathrm{d}s=\int_0^C G^K(t,s)\bm H^\top(\bm{X}(s))\mathrm{d}s+\;\;\Phi_K^\top(t)\int_0^C\bm{v}_K(s)\bm H^{\top}(\bm{X}(s))\mathrm{d}s.
    \end{align*}
Substituting into the definition of $\tilde{\bm\Theta}_i$ in Theorem \ref{thm-optimizer}, we obtain:
\begin{align*}
    \tilde{\bm\Theta}_i(\bm X(t))
    =\bm\Theta_i(\bm X(t))+\left(\bm\Phi_{K-1}^\top(t)\int_0^C\bm{v}_{K-1}(s)X_i(s)\mathrm{d}s,\right.\ldots,&\\
\left.\bm\Phi_1^\top(t)\int_0^C\bm{v}_1(s)X_i(s)\mathrm{d}s,\bm\Phi_K^\top(t)\int_0^C\bm{v}_K(s)\bm H^{\top}(\bm{X}(s))\mathrm{d}s\right).&
\end{align*}

Let $\bm\theta_i=(-\omega_{i1},\ldots,-\omega_{i(K-1)},\bm\beta_i^\top)^\top$, and $\bm\alpha_i=(\alpha_1,\ldots,\alpha_K)$. Consider a transformed coefficient vector $\tilde{\bm\alpha}_i=(\tilde{\alpha}_{i1},\ldots,\tilde{\alpha}_{iK})$ defined by
\begin{align}
    \label{alpha-trans}\tilde{\alpha}_{ik}=\alpha_{ik}+\sum_{l=k}^{K-1}\omega_{il}\int_0^C\psi_{lk}(s)X_i(s)\mathrm d s-&\\
    \nonumber
    \int_0^C\psi_{Kk}(s)\bm H^\top(\bm X(s))\bm\beta_i\mathrm ds, \quad k=1,\dots,K,&
\end{align}
where the first summation term $\sum_{l=k}^{K-1}\omega_{il}\int_0^C\psi_{lk}(s)X_i(s)\mathrm d s$ is omitted when $k=K$.

With this transformation, we have
$$\tilde{\mathcal{F}}^i_{\bm\theta_i, \tilde{\bm\alpha}_i}(\bm X, t)=\mathcal{F}^i_{\bm\theta_i, \bm\alpha_i}(\bm X, t),$$
and consequently,
$$\tilde{\mathcal{L}}_i( \bm\theta_i ,\tilde{\bm\alpha}_i\mid \bm{X}) =\mathcal L_i(\bm\theta_i,\bm\alpha_i\mid\bm X)$$
for any fixed $\bm\theta_i$ if $\bm\alpha_i$ and $\tilde{\bm\alpha}_i$ satisfy \eqref{alpha-trans}. 

Define the optimal values $$M_1:=\min_{\bm\theta_i,\bm\alpha_i}\mathcal L_i(\bm\theta_i,\bm\alpha_i\mid\bm X),\quad M_2:=\min_{{\bm\theta}_i ,{\bm\alpha}_i}\tilde{\mathcal{L}}_i( {\bm\theta}_i ,{\bm\alpha}_i\mid \bm{X})$$

We now prove $M_1=M_2$. 

For any $\epsilon>0$, there exists $(\bm\theta_i^{\epsilon},\bm\alpha_i^{\epsilon})$ such that
$$\mathcal L_i(\bm\theta_i^{\epsilon},\bm\alpha_i^{\epsilon}\mid\bm X)<M_1+\epsilon,$$
Define $\tilde{\bm\alpha}_i$ using \eqref{alpha-trans} with $\bm\alpha_i=\bm\alpha_i^{\epsilon}$ and $\bm\theta_i=\bm\theta_i^{\epsilon}$. Then 
$$M_2\leq\tilde{\mathcal{L}}_i(\bm\theta_i^{\epsilon},\tilde{\bm\alpha}_i\mid\bm X)=\mathcal L_i(\bm\theta_i^{\epsilon},\bm\alpha_i^{\epsilon}\mid\bm X)<M_1+\epsilon.$$
Letting $\epsilon\to0$ yields $M_2\leq M_1$. Similarly, using the inverse transformation in equation \eqref{alpha-trans-inverse}, one can show $M_1\leq M_2$. Therefore, we conclude that \( M_1 = M_2 \).

Now, if $(\hat{\bm\theta_i},\hat{\bm\alpha_i})$ minimizes $\mathcal L_i(\bm\theta_i,\bm\alpha_i\mid\bm X)$, define $\tilde{\bm\alpha}_i$ using \eqref{alpha-trans} with $\bm\alpha_i=\hat{\bm\alpha_i},\bm\theta_i=\hat{\bm\theta_i}$. Then 
\begin{align*}
\tilde{\mathcal{L}}_i(\hat{\bm\theta_i},\tilde{\bm\alpha}_i\mid\bm X)=M_1=M_2,
\end{align*}
so $(\hat{\bm\theta_i},\tilde{\bm\alpha}_i)$ minimizes $\tilde{\mathcal L}_i(\bm\theta_i,\bm\alpha_i\mid\bm X)$. 

Conversely, if \((\tilde{\bm\theta}_i,\tilde{\bm\alpha}_i)\)  minimizes $\tilde{\mathcal L}_i(\bm\theta_i,\bm\alpha_i\mid\bm X)$, let $\tilde{\bm\theta}_i=(-\tilde\omega_{i1},\ldots,-\tilde\omega_{i(K-1)},\tilde{\bm\beta}_i^\top)^\top$ and define $\hat{\bm\alpha}_i=(\hat{\alpha}_{i1},\ldots,\hat{\alpha}_{iK})$ via the inverse transformation:
\begin{align}
    \label{alpha-trans-inverse}\hat{\alpha}_{ik}=\tilde{\alpha}_{ik}-\sum_{l=k}^{K-1}\tilde\omega_{il}\int_0^C\psi_{lk}(s)X_i(s)\mathrm d s+\int_0^C\psi_{Kk}(s)\bm H^\top(\bm X(s))\tilde{\bm\beta}_i\mathrm ds,\quad k=1,\dots,K.
\end{align}
Then \((\tilde{\bm\theta}_i,\hat{\bm\alpha}_i)\) minimizes  $\mathcal L_i(\bm\theta_i,\bm\alpha_i\mid\bm X)$.

By definition, there exists an identity map $\mathrm{id}$ and the bijection $\mathcal{M}$ constructed via equation~\eqref{alpha-trans-inverse} such that
\[
\operatorname{id}(\tilde{\Theta}) = \hat{\Theta}, \quad \text{and} \quad
\mathcal{M}(\tilde{\mathcal{A}}) = \hat{\mathcal{A}}.
\]
\end{proof}

\section{Supporting Results}
\label{app-edge}
\subsection{Nonlinear Pendulum with Misspecified Library}\label{sec_ap:NP}
For the pendulum system, we further consider the governing equation:
\[
\frac{\mathrm{d}^2X(t)}{\mathrm{d}t^2} = -\sin(X(t)) - 0.2\,\sin(X(t))\cos(X(t)).
\]
This model introduces an additional nonlinear interaction term, \( \sin(X)\cos(X) \), which is not included in the candidate function library
\(
\bm{H}(X) = \left(1,\, X,\, X^2,\, X^3,\, X^4,\, \sin(X),\, \cos(X)\right)^\top.
\)

\begin{figure}[htbp]
    \centering
    \includegraphics[width=.95\linewidth]{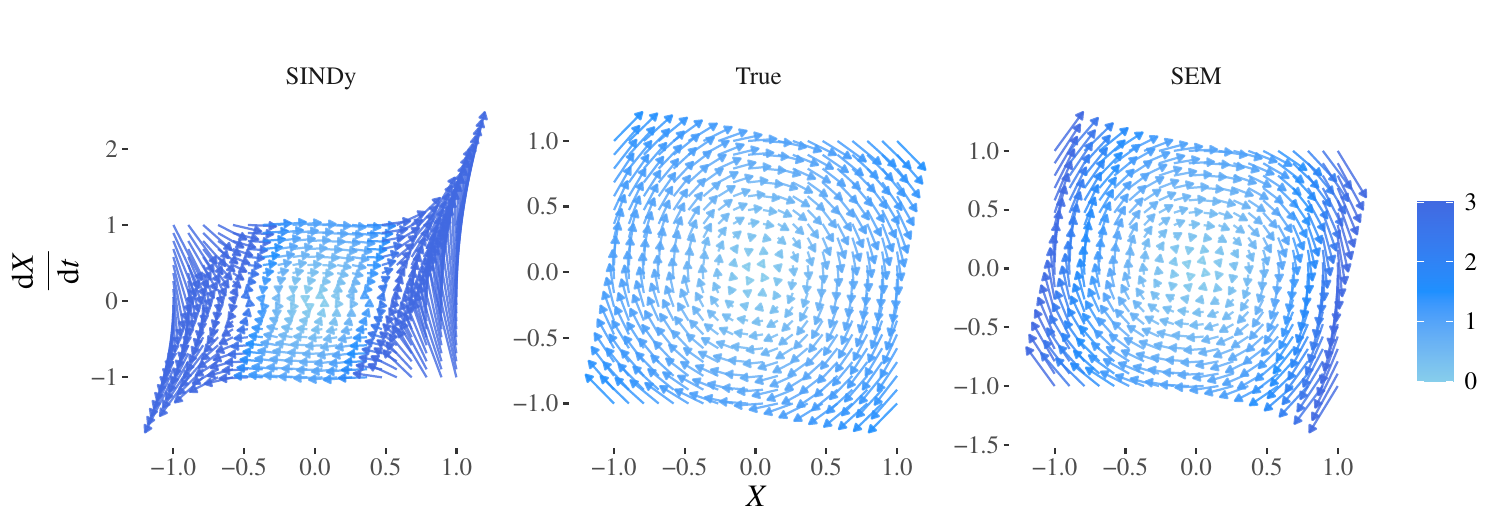}
    \caption{\small 
    True and average estimated vector fields for the pendulum system with nonlinear damping under the setting \( n = 150 \) and \( \gamma = 0.05 \). 
    The color indicates the magnitude of the velocity vectors.
    }
    \label{fig:vf_pen_damped}
\end{figure}

Figure~\ref{fig:vf_pen_damped} displays the true and average estimated vector fields obtained from 2000 simulation runs under the same setting (\( n = 150 \), \( \gamma = 0.05 \)). 
Even with the misspecified candidate library, SEM continues to outperform gradient-based approaches such as SINDy, producing vector fields that are much closer to the true dynamics.

\

\clearpage
\subsection{Directed Brain Networks and Centrality Measures}
\begin{figure}[htbp]
    \centering
    \includegraphics[width=0.85\linewidth]{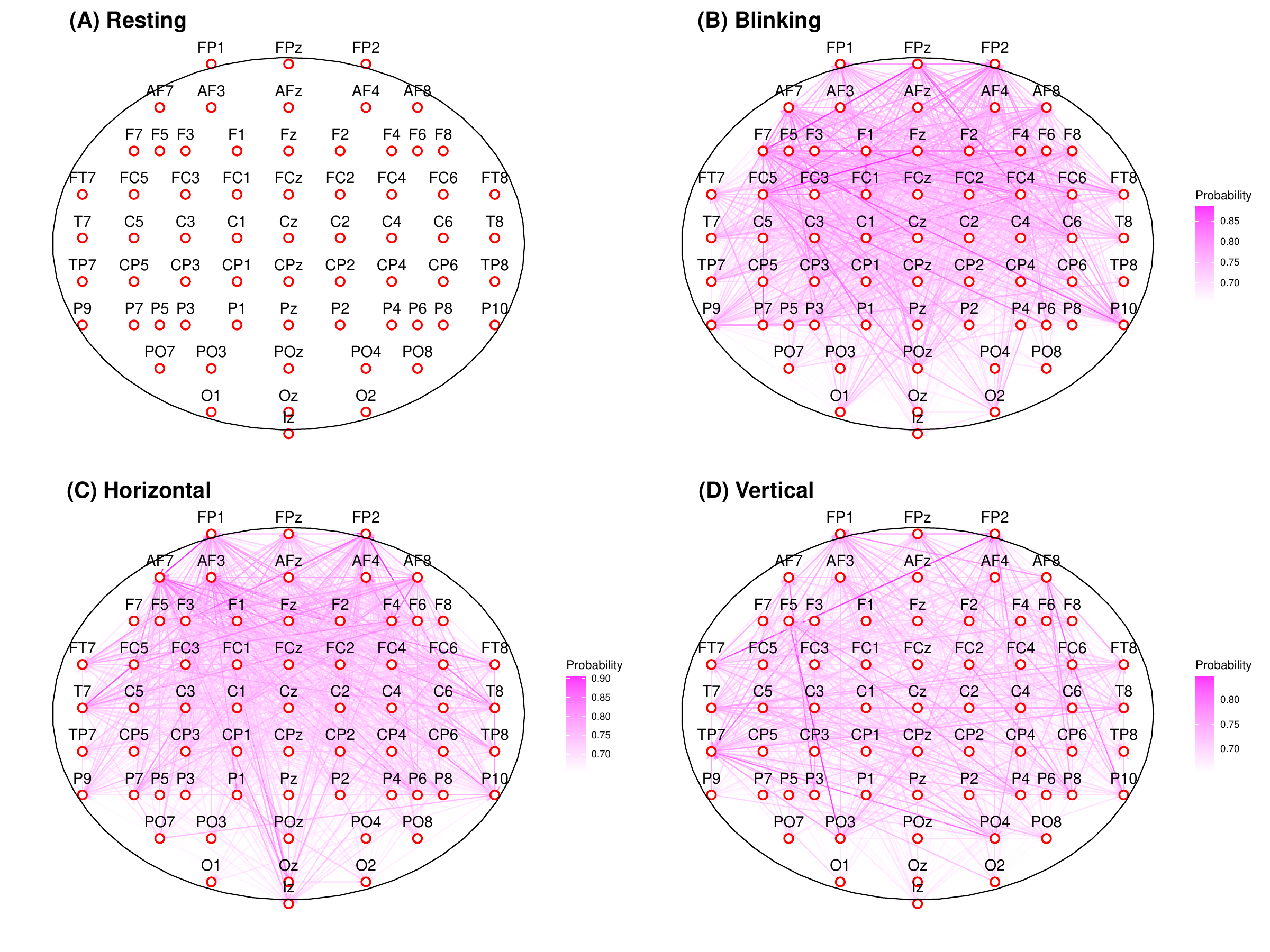}
    \caption{\small Directed brain networks for different tasks: (A) resting state,  (B) eye-blinking, (C) horizontal eye movement,  (D) vertical eye movement. Edge color indicates the empirical probability of each edge occurring across participants.
    }
    \label{fig:all-tasks}
\end{figure}

\begin{figure}[htbp]
    \centering
    \includegraphics[width=0.9\linewidth]{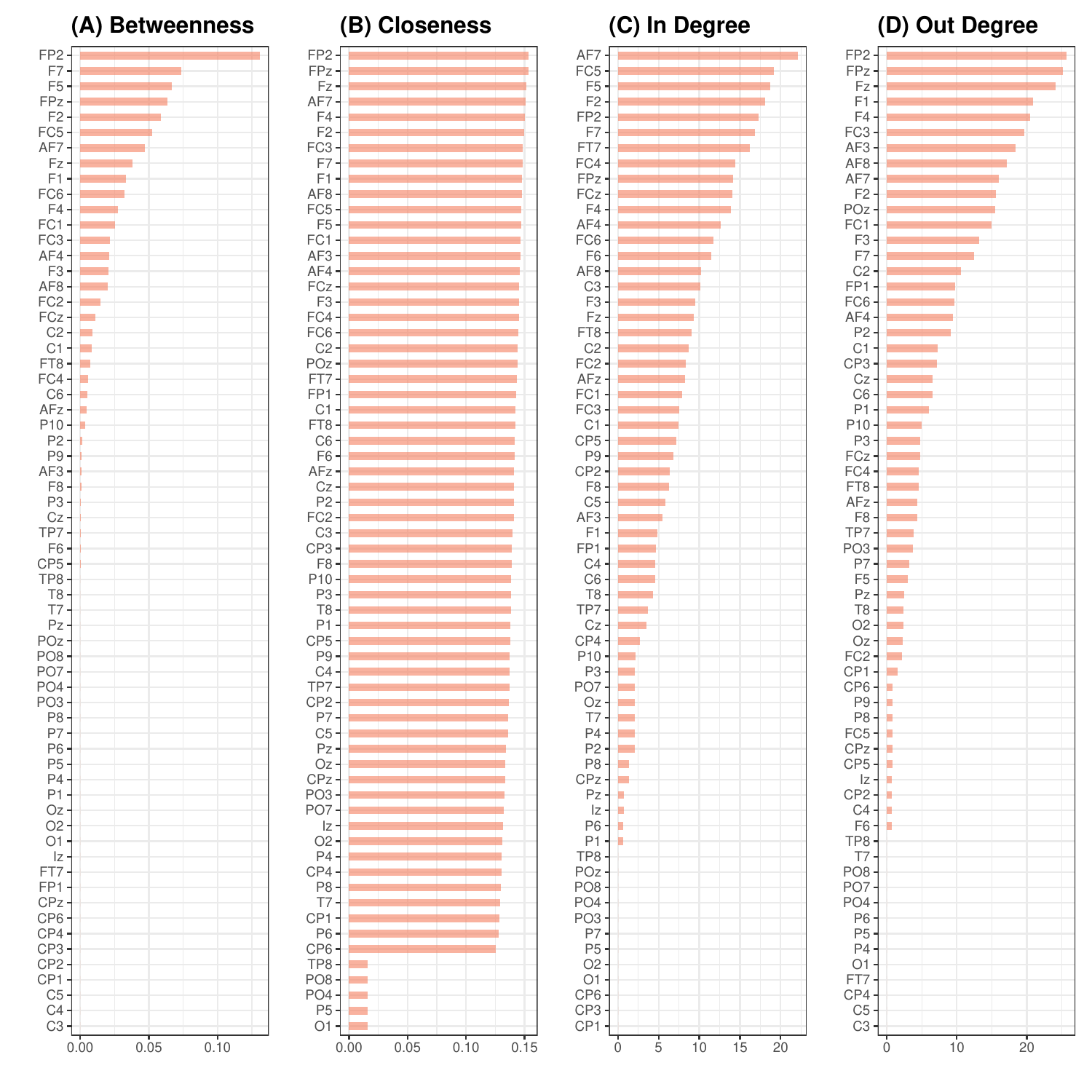}
    \caption{\small A comparison of centrality measures. The diagrams display rank-ordered distributions of betweenness centrality, closeness centrality, in-degree, and out-degree for the network associated with the eye-blinking task, as shown in Figure~\ref{fig:net_tasks}.}
    \label{fig:graph-count-blink}
\end{figure}

\begin{figure}
    \centering
    \includegraphics[width=0.9\linewidth]{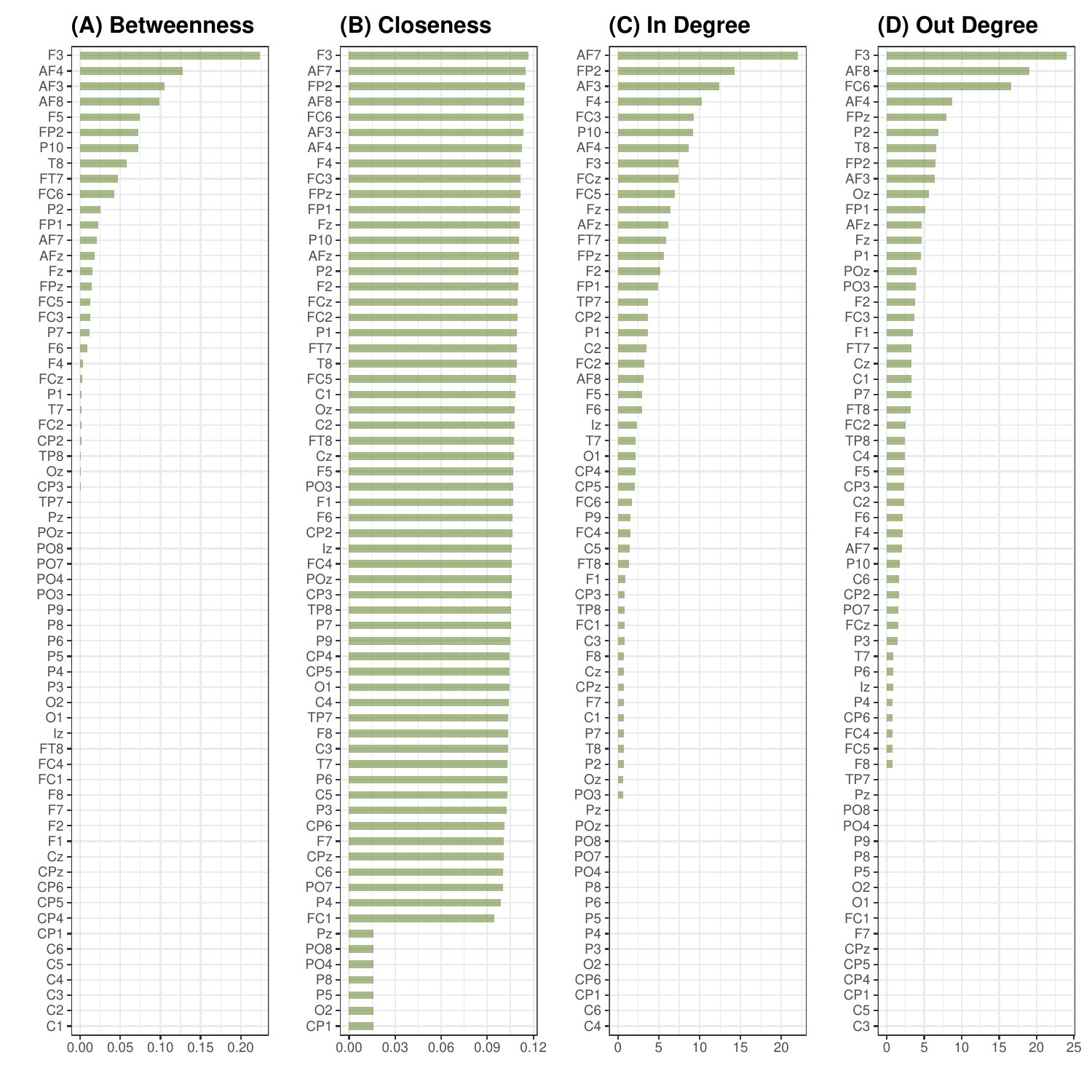}
   \caption{\small A comparison of centrality measures. Diagrams show rank-ordered distributions of betweenness, closeness, in degree and out degree for the network of horizontal eyeball movement task shown in Figure \ref{fig:net_tasks}}
  \label{fig:graph-count-left}
\end{figure}

\begin{figure}[htbp]
    \centering
    \includegraphics[width=0.9\linewidth]{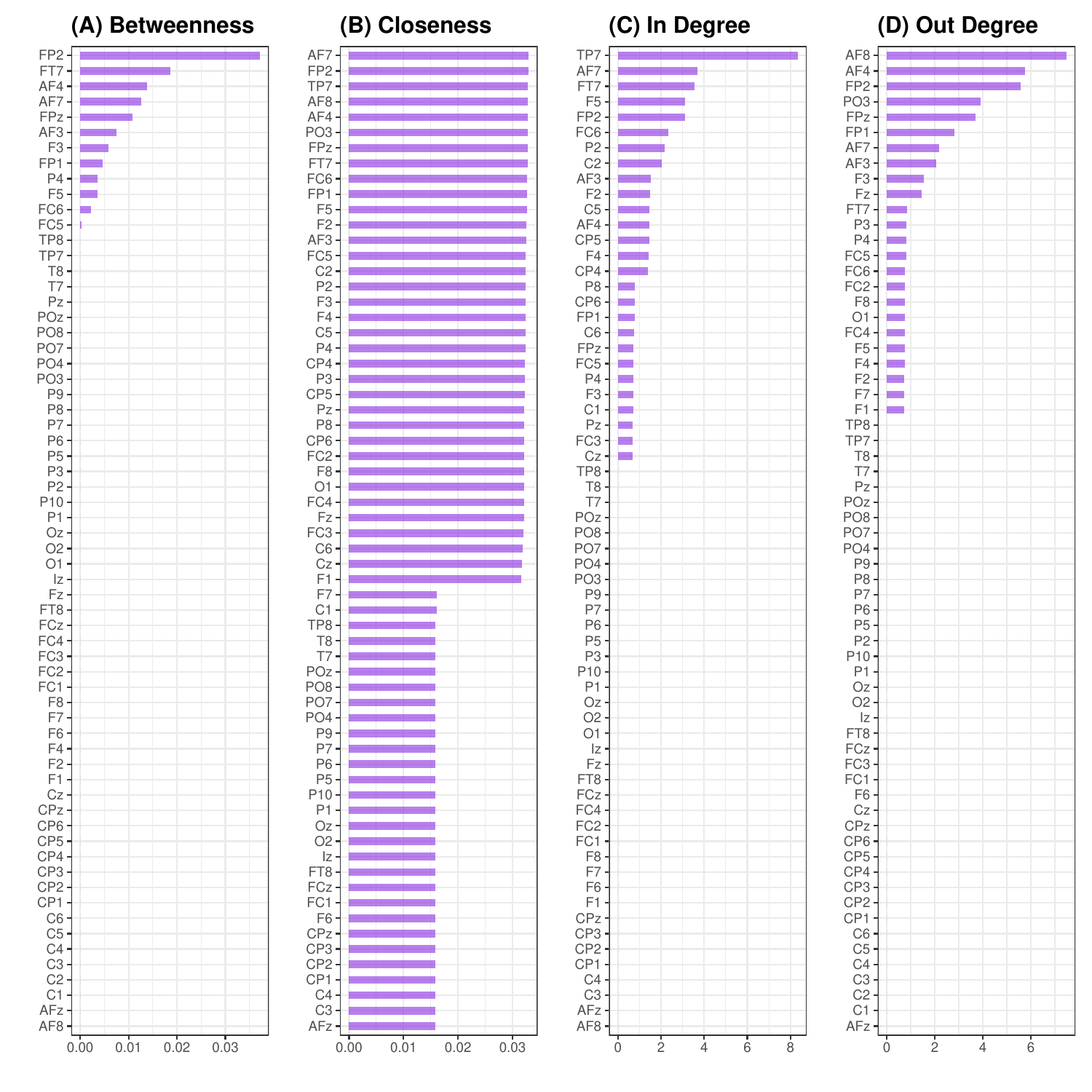}
    \caption{\small A comparison of centrality measures. Diagrams show rank-ordered distributions of betweenness, closeness, in degree and out degree for the network of vertical eyeball movement task shown in Figure \ref{fig:net_tasks}}
    \label{fig:graph-count-up}
\end{figure}

\end{document}